\documentclass{article}

     \PassOptionsToPackage{numbers, compress}{natbib}
 \usepackage[preprint]{neurips_2026}

\usepackage[utf8]{inputenc} 
\usepackage[T1]{fontenc}    
\usepackage{hyperref}       
\usepackage{url}            
\usepackage{booktabs}       
\usepackage{amsfonts}       
\usepackage{nicefrac}       
\usepackage{microtype}      
\usepackage{xcolor}         
\usepackage{algpseudocode}
\usepackage{algorithm}
\usepackage{lipsum}
\usepackage{wrapfig}
\usepackage{mathtools}
\usepackage{amsthm}
\usepackage{amsmath,amsfonts,amssymb,color}
\usepackage{mathtools}
\usepackage{dsfont}
\usepackage{epsfig}
\usepackage{epstopdf}
\usepackage{pifont}
\usepackage{caption}
\usepackage{subcaption}

\usepackage[utf8]{inputenc} 
\usepackage[T1]{fontenc}    
\usepackage{hyperref}       
\usepackage{url}            
\usepackage{booktabs}       
\usepackage{amsfonts}       
\usepackage{nicefrac}       
\usepackage{microtype}      
\usepackage{xcolor}         

\title{Transformers as In-Context Samplers: From Closed-Form Diffusion to Estimation-Free Sampling}
\usepackage[utf8]{inputenc} 
\usepackage[T1]{fontenc}    
\usepackage{hyperref}       
\usepackage{url}            
\usepackage{booktabs}       
\usepackage{amsfonts}       
\usepackage{nicefrac}       
\usepackage{microtype}      
\usepackage{xcolor}         
\usepackage{algpseudocode}
\usepackage{algorithm}
\usepackage{lipsum}
\usepackage{wrapfig}
\usepackage{mathtools}
\usepackage{amsthm}
\usepackage{amsmath,amsfonts,amssymb,color}
\usepackage{mathtools}
\usepackage{dsfont}
\usepackage{epsfig}
\usepackage{epstopdf}
\usepackage{pifont}

\newtheorem{theorem}{Theorem}

\newtheorem{remark}{Remark}

\DeclarePairedDelimiterX{\norm}[1]{\lVert}{\rVert}{#1}

\author{%
  Arman Adibi$^{1}$ \quad
  Alireza Jafari$^{2}$ \quad
  Mohammad Ghavamzadeh$^{3}$ \quad
  Hadi Daneshmand$^{2}$ \\[3pt]
  $^{1}$School of Computer and Cyber Sciences, Augusta University \\
  $^{2}$Department of Computer Science, University of Virginia \\
  $^{3}$Qualcomm AI Research \\[3pt]
  \texttt{aadibi@augusta.edu}
}

\begin{document}

\maketitle

\begin{abstract}
A growing body of work establishes that large language models are not mere statistical memorizers, but are capable of in-context learning: performing inference at test time using only examples provided in the prompt, without any parameter updates. Prior theoretical work has shown that this capability extends to supervised learning tasks such as linear regression. We prove that in-context learning extends further to \emph{data generation}: frozen transformers can simulate iterative generative samplers from in-context samples. We first show that transformers can realize closed-form and smoothed closed-form diffusion samplers. The construction identifies a concrete generative role for softmax attention: it computes responsibility weights and weighted empirical averages, while feedforward layers implement Euler updates.

To empirically relate these constructions to pretrained language models, we study \emph{semantic-topic sampling}: prompts consisting of words drawn from a common semantic category, such as animals, foods, or cities. Across transformer layers, the normalized hidden states exhibit a two-stage geometry: they move toward a uniform spherical reference in intermediate layers and then return to structured, topic-dependent representations near the output. We further measure an interacting-particle energy on these hidden-state clouds and observe the same U-shaped pattern.
We then prove that transformers can approximate a finite forward--backward particle sampler inspired by estimation-free sampling. This expressivity result motivates an analogy with the observed geometry without implying a U-shaped hidden-state energy trajectory. 
\end{abstract}
\section{Introduction}

"In-context learning" \cite{brown2020language} has changed how we think about transformers. A transformer can adapt its behavior at inference time using only information provided in a prompt, without updating its parameters. This phenomenon was first made prominent by large language models, where a prompt containing a few demonstrations can induce the model to perform a new task~\cite{brown2020language,chowdhery2023palm,achiam2023gpt,touvron2023llama,touvron2023llama2}. It has since become a central paradigm for studying how models use demonstrations, instructions, and contextual information at inference time~\cite{dong2024survey}.

Existing studies have mostly focused on conditional data generation, specifically for supervised learning: the prompt contains examples from an unknown task, and the model predicts the output for a new query. A standard formalization studies prompts of the form
\[
\underbrace{(x_1,f(x_1),\ldots,x_k,f(x_k),x_{\mathrm{query}})}_{\text{prompt}}
\quad\longmapsto\quad
\underbrace{f(x_{\mathrm{query}})}_{\text{completion}} .
\]
For example, Garg et al.~\cite{garg2022what} ask whether transformers can learn simple function classes in context, such as linear functions, sparse linear functions, decision trees, and shallow neural networks. Subsequent work studies whether transformers implement particular learning algorithms in context, including ridge regression, gradient descent, higher-order optimization, and algorithm selection~\cite{akyurek2023what,li2023transformers,bai2023transformers,fu2023transformers}. These results suggest that transformers can do more than match surface patterns in the prompt: they can execute nontrivial computations inside their forward pass.

Transformers were originally proposed for conditional text generation~\cite{vaswani2017attention}, 
but have since become the backbone of powerful unconditional generative models, including 
latent diffusion models~\cite{rombach2022high}. While recent studies have extensively 
investigated the mechanistic interpretation of in-context learning in conditional text generation, 
it remains an open question whether such capabilities extend to unconditional data generation 
and broader generative modeling applications. In this paper, we ask whether in-context learning extends to data generation:
\begin{center}
\emph{Can a transformer implement a sampler in context?}
\end{center}
Given i.i.d. samples $x_1, \dots, x_n \in \mathbb{R}^d$ from an unknown distribution $\mathcal{P}$, a sampling algorithm aims to generate a fresh sample $x_{n+1} \neq x_1, \dots, x_n$ from $\mathcal{P}$. Akin to the standard settings of generative models, we assume that the density function of the data distribution is unknown, and we only have access to i.i.d. samples from the target distribution. 
Inspired by prior in-context learning studies~\cite{ahn2023transformers,garg2022what}, we encode the sampling problem as
\[
\text{input:} \quad x_1, \dots, x_n\sim_{\text{i.i.d.}} \mathcal{P}
\quad\longmapsto\quad \text{output:} \quad x_{n+1} \sim \mathcal{P},
\]
where $x_1, \dots, x_n$ are encoded in the word embeddings of a large language model and $x_{n+1}$ is an output embedding vector.
We refer to this prompting strategy as \emph{in-context sampling}: the transformer parameters 
remain fixed while the distribution of $\mathcal{P}$ changes at test-time. A model can leverage two sources 
of information for inference: statistics from training data and information from in-context 
samples. In-context learning emerges when the model relies on the latter, adapting its response 
to the in-context input at test time. 

\begin{figure}[h!]
    \centering
    \includegraphics[width=0.8\textwidth]{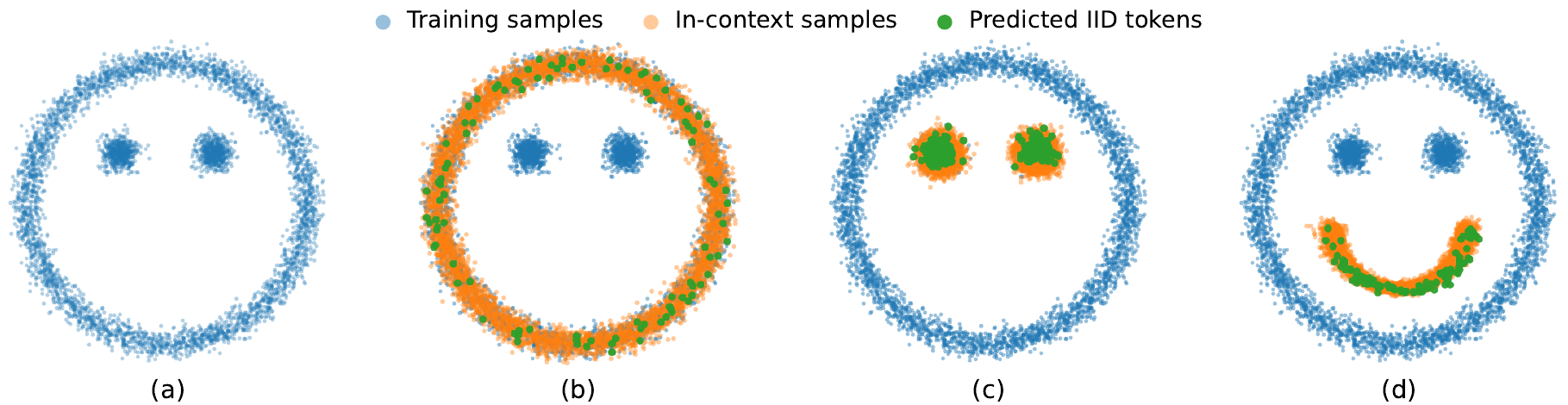} 
    \caption{
\footnotesize{
\textbf{In-context sampling.}
We train a transformer to generate points from the blue training distribution in panel~(a), which contains the face boundary and eyes but excludes the smiling mouth.
At test time, the model receives in-context samples shown in orange and generates new samples shown in green.
When the context samples are drawn from components present during training, the generated samples follow the face boundary in panel~(b) and the eyes in panel~(c).
Panel~(d) gives the out-of-distribution test: the context samples form a smiling mouth, a component entirely absent from training.
The model nevertheless generates new samples along the same unseen curve.
This indicates that the transformer is using in-context samples for inference, rather than relying on statistics of the training distribution.
See the Experiments section for training and data generation details.}}
    \label{fig:in-context-sampling}
\end{figure}

Figure~\ref{fig:in-context-sampling} illustrates the phenomenon that motivates our study. 
The model is trained on samples from a face distribution containing the face boundary and eyes, but no smiling mouth, where samples $x_1, \dots, x_n$ are encoded in word embeddings. 
When the in-context samples come from components seen during training, such as the boundary or the eyes, the generated samples follow the corresponding prompt distribution. 
The key test is the smiling mouth: although this component is absent from training, providing mouth-shaped samples in context causes the transformer to generate new samples along the same unseen curve. 
Thus, the model is not simply reproducing the global training distribution or selecting among memorized components. 
Rather, the inference at test time is based on in-context samples.  
Our goal is to explain the mechanism underlying this capability.

\subsection*{Contributions}

We make the following contributions to demonstrate the generative power of transformers.

\emph{In-context simulation of diffusion models.} We prove that transformers can simulate 
closed-form diffusion models~\cite{scarvelis2025closedform} from in-context samples. 
Closed-form diffusion models leverage the explicit score function to simulate the backward 
diffusion process. We show that softmax attention heads are particularly well-suited to 
compute the empirical score function directly from in-context samples.

\emph{Mechanistic analysis of pretrained LLMs.} We examine the data generation mechanism in pretrained language models, including 
GPT-2~\cite{radford2019language}, 
Llama-3.3-70B-Instruct~\cite{grattafiori2024llama}, 
OpenLM-Llama-13B/OpenLLaMA~\cite{openlm2023openllama}, 
Cerebras-GPT-13B~\cite{dey2023cerebras}, 
Qwen2.5~\cite{qwen2024qwen2}, 
Falcon-40B~\cite{almazrouei2023falcon}, 
OPT-66B~\cite{zhang2022opt}, 
and BLOOM-7B1~\cite{workshop2022bloom}.
The models are prompted by distinct words sampled without replacement from the same semantic topic. This gives a controlled semantic context, although the words are not independent draws. Tracking the empirical embedding distributions across layers, we observe smaller discrepancies to a uniform spherical reference in middle layers and larger discrepancies near the output. The interaction energy of the word embeddings exhibits a similar U-shaped pattern.

\emph{Estimation-free sampling.} We prove that transformers can approximate a finite forward--backward particle computation inspired by estimation-free sampling (EFS). The construction first simulates particle gradient descent and then transports a reference point using an explicit reverse-time discretization. This two-stage mechanism provides a qualitative analogy with the distribution shifts observed in pretrained transformers; the theorem establishes approximation of the terminal state.

\paragraph{Connections to prior work.}
Our work builds on the algorithmic view of in-context learning, where transformers are studied as models that can execute non-trivial computations from prompts rather than only predict labels~\cite{garg2022what,akyurek2023what,ahn2023transformers,li2023transformers,bai2023transformers}. It is also related to diffusion and score-based generative modeling~\cite{ho2020denoising,song2020score,song2020denoising}, as well as Estimation-Free Sampling~\cite{daneshmand2025data}. The closest work connecting diffusion and in-context learning is Prompt Diffusion~\cite{wang2023promptdiffusion}, which trains a diffusion model to perform visual tasks from in-context examples. Our direction is different: we ask whether a frozen transformer can itself implement diffusion-style and particle-based sampling algorithms from the prompt. A more detailed discussion of related work is deferred to Appendix~\ref{app:related}.


\section{Preliminaries}
\subsection{Closed-form diffusion process}
\label{subsec:closed-form-diffusion}

Score-based diffusion models~\cite{song2020score} are among the most powerful generative models for transporting a simple reference distribution, such as a Gaussian distribution, to a target data distribution. In these models, the transport dynamics are driven by the score function, i.e., the gradient of the log-density of a smoothed data distribution. In practice, this score is typically approximated by a neural network trained through denoising.

In continuous time, the reverse-time generative dynamics can be described through Anderson's reverse-time diffusion framework~\cite{anderson1982reverse}. In an abstract form, the reverse time process is written as
\begin{equation}
    d z_t
    =
    \left(
    -z_t-\nabla_z \log p_t(z_t)
    \right)dt
    +
    dW_t,
    \label{eq:reverse-diffusion-abstract}
\end{equation}
where \(p_t\) denotes the density of the target distribution after Gaussian smoothing at time \(t\), and \(W_t\) is a Wiener process. The term \(\nabla_z \log p_t(z_t)\) drives the sample toward regions of high probability under the smoothed data distribution.~\footnote{The negative sign of $\nabla \log p_t(z_t)$ arises because the process runs in reverse time, from $T$ back to $0$.} The Gaussian smoothing and stochastic noise injection play an annealing role: at large noise levels, the density is smoother and easier to explore, while at smaller noise levels, the dynamics refine samples toward the data distribution.

Closed-form diffusion models use the observation that, for a finite empirical distribution, the Gaussian-smoothed density can be written explicitly as a Gaussian mixture~\cite{scarvelis2025closedform}. Let $
\{x_i\}_{i=1}^N\subset\mathbb{R}^d$ be the empirical dataset. For \(t\in(0,1)\), define
\[
\rho_t^\star(z)
=
\frac{1}{N}
\sum_{i=1}^N
\phi\!\left(z;t x_i,(1-t)^2 I_d\right),
\qquad z\in\mathbb{R}^d,
\]
where \(\phi(\cdot;\mu,\Sigma)\) is the Gaussian density with mean \(\mu\) and covariance \(\Sigma\). Thus, \(\rho_t^\star\) is a mixture of isotropic Gaussians centered at the scaled samples \(t x_i\), with covariance \((1-t)^2 I_d\).

For a current point \(z\), define the responsibility weight of sample \(x_i\) by
\begin{equation} \label{eq:resp}
w_i(t,z)
=
\frac{
\phi\!\left(z;t x_i,(1-t)^2 I_d\right)
}{
\sum_{j=1}^N
\phi\!\left(z;t x_j,(1-t)^2 I_d\right)
},
\qquad i=1,\ldots,N.
\end{equation}
These weights satisfy \(w_i(t,z)\ge 0\) and \(\sum_{i=1}^N w_i(t,z)=1\). They measure how much the Gaussian component centered at \(t x_i\) contributes to the density at \(z\). Define the corresponding weighted mean
\(
k_t(z)
=
\sum_{i=1}^N w_i(t,z)(t x_i).
\)
Since the distribution of $p_t^*(z)$ is a Gaussian mixture with explicit density function, its score can be computed in closed-form as \cite{scarvelis2025closedform}
\begin{equation}
    \nabla_z \log \rho_t^\star(z)
    =
    \frac{1}{(1-t)^2}
    \bigl(k_t(z)-z\bigr).
    \label{eq:closed-form-score}
\end{equation}
Thus, the score tells the sampler how to move \(z\): it points from the current point \(z\) toward the weighted average \(k_t(z)\) of the data samples.

The sampler converts this score into an update direction, which we write it as
\begin{equation}
    v_t(z)
    =
    \frac{1}{t}
    \left(
    z+(1-t)\nabla_z\log \rho_t^\star(z)
    \right).
    \label{eq:cfd-velocity-definition}
\end{equation}
Here \(v_t(z)\in\mathbb{R}^d\) is the direction in which the sampler moves the current point \(z\) at time \(t\). Substituting~\eqref{eq:closed-form-score} gives
\begin{equation}
    v_t(z)
    =
    \frac{1}{t}
    \left(
    z+\frac{1}{1-t}(k_t(z)-z)
    \right) \nonumber
    =
    -\frac{1}{1-t}z
    +
    \frac{1}{t(1-t)}k_t(z).
    \label{eq:cfd-velocity}
\end{equation}
Therefore, the main data-dependent computation is the weighted mean \(k_t(z)\). Once \(k_t(z)\) is known, the update direction \(v_t(z)\) is a fixed linear combination of \(z\) and \(k_t(z)\).

Given a time grid \(\tau=\{(t_s,h_s)\}_{s=0}^{S-1}\), where \(0<t_s<1\) and \(h_s>0\), the closed-form diffusion sampler relies on the following recurrence for sampling:
\begin{equation}
    z_{s+1}
    =
    z_s+h_s v_{t_s}(z_s),
    \qquad
    s=0,\ldots,S-1.
    \label{eq:cfd-euler}
\end{equation}

This explicit form is central to our construction. The sampler does not require a trained score network: the score and update direction are computed directly from the in-context samples \(x_1,\ldots,x_N\). In later sections, we show that a frozen transformer can implement this computation using attention to compute the weights \(w_i(t,z)\) and the weighted mean \(k_t(z)\), and using feedforward layers to perform the Euler update.

\paragraph{Smoothing.}
\label{subsec:smoothed-closed-form-diffusion}

Since the closed-form diffusion model may memorize data, \cite{scarvelis2025closedform} proposes smoothing to avoid data memorization. Smoothing is obtained by averaging the closed-form mean over fixed perturbations of the current state. Let \(\sigma\ge 0\) and \(\mathcal{E}=\{\epsilon_m\}_{m=1}^M\subset\mathbb{R}^d\) be fixed perturbation vectors. Define \(k_{\sigma,t}(z)=M^{-1}\sum_{m=1}^M k_t(z+\sigma\epsilon_m)\), where \(k_t(\cdot)\) is the closed-form responsibility-weighted mean defined above. The smoothed velocity field is
\begin{equation}
    v_{\sigma,t}(z)
    =
    -\frac{1}{1-t}z
    +
    \frac{1}{t(1-t)}k_{\sigma,t}(z).
    \label{eq:smoothed-cfd-velocity}
\end{equation}
Thus, the smoothed sampler replaces \(k_t(z)\) by the averaged mean \(k_{\sigma,t}(z)\). Setting \(\sigma=0\) recovers the closed-form diffusion method. \cite{scarvelis2025closedform} has examined the data generation capability of the above method in comparison to original denoising diffusion models.   

This modification preserves the same transformer-realization structure: attention computes \(k_t(z+\sigma\epsilon_m)\) for each fixed perturbation, the heads are averaged to obtain \(k_{\sigma,t}(z)\), and the feedforward layer applies the affine update~\eqref{eq:smoothed-cfd-velocity}.
\subsection{Embedding encoding}
\label{subsec:embedding-encoding}

We encode a closed-form diffusion instance in the input word embedding matrix of a transformer, denoted by
\[
X
=
\mathrm{Enc}(\{x_i\}_{i=1}^N,z_0)
\in \mathbb{R}^{N\times p},
\qquad
p=3d+2,
\]
where  \(z_0\in\mathbb{R}^d\) is a Gaussian random vector which is equivalent to starting state of the closed-form diffusion model defined in~\eqref{eq:smoothed-cfd-velocity}. The encoded embedding consists of \(N\) data tokens; the final token also stores the sampler state. 
We write \(X\in\mathbb{R}^{N\times p}\), with \(p=3d+2\), and define its rows by
\[
X_i =
\bigl[x_i^\top,\ \|x_i\|^2,\ 0_d^\top,\ 0_d^\top,\ 1\bigr],
\qquad i=1,\ldots,N-1,\quad
X_{N}
=
\bigl[x_N^\top,\ \|x_N\|^2,\ z_0^\top,\ 0_d^\top,\ 1\bigr].
\]
Each row stores one in-context sample, and the final row also stores the initial sampler state \(z_0\).
Equivalently, each row is partitioned as
\(
X_i
=
\bigl[
X_i^{(x)}
\mid
X_i^{(r)}
\mid
X_i^{(z)}
\mid
X_i^{(k)}
\mid
X_i^{(1)}
\bigr],
\)
where \(X_i^{(x)},X_i^{(z)},X_i^{(k)}\in\mathbb{R}^d\), \(X_i^{(r)}\in\mathbb{R}\), and \(X_i^{(1)}\in\mathbb{R}\).  These four blocks serve distinct purposes in the design: the $x$ and $r$ blocks encode training samples for in-context learning; the $z$ block acts as memory for newly generated samples; and the $k$ block functions as a scratchpad for intermediate computation, inspired by prior work~\cite{daneshmand2026context}.
The state readout map $\Pi_{\text{state}}: \mathbb{R}^{m\times p} \to \mathbb{R}^d$ extracts the \(z\)-block of the final row:
$
\Pi_{\mathrm{state}}(X)
=
X_{N}^{(z)}
=
X_{N,\;d+2:\,2d+1}
$. We will later use the above notation to reconstruct the output of a transformer.\subsection{Transformer}
\label{subsec:transformer}

We use residual softmax self-attention and positionwise feedforward blocks~\cite{vaswani2017attention}, with no layer normalization or dropout in the theoretical constructions. Each block has the form \(Y=X+\mathrm{Attn}(X)\), \(X^+=Y+\mathrm{FF}(Y)\), where \(\mathrm{FF}\) is a shared, two-layer ReLU network with a freely chosen finite hidden width. Attention is full self-attention; the closed-form diffusion construction also permits standard causal attention. The usual attention scaling is absorbed into the query weights. Let \(\mathcal{T}_\theta:\mathbb{R}^{m\times p}\to\mathbb{R}^{m\times p}\) denote a depth-\(L\) transformer, where \(m\) is the number of tokens, \(p\) is the token dimension, \(L\in\mathbb{N}\) is the number of layers, and \(\theta\) denotes the collection of attention and feedforward weight matrices. Thus, the transformer maps an input embedding matrix to an output embedding matrix of the same size.

We generate a sample by applying the state readout map \(\Pi_{\mathrm{state}}:\mathbb{R}^{m\times p}\to\mathbb{R}^d\), defined in the previous section, to the final embedding matrix. For a sampling instance \(\{x_i\}_{i=1}^N\) with initial state \(z_0\), the transformer output is
\(
\widehat z_L
=
\Pi_{\mathrm{state}}\!\left(
\mathcal{T}_\theta\bigl(\mathrm{Enc}(\{x_i\}_{i=1}^N,z_0)\bigr)
\right).
\)
Our goal is to construct parameter choices \(\theta\) such that \(\widehat z_L\) matches the output of a specified iterative generative sampler.
\section{Simulation of Closed-form Diffusion with Transformers}
\label{sec:cfd-transformer-simulation}

We now show that the transformer architecture in Section~\ref{subsec:transformer} can simulate the (smoothed) closed-form diffusion sampler by a suitable choice of its parameters. The prompt contains the empirical samples \(\{x_i\}_{i=1}^N\subset\mathbb{R}^d\) and the initial state \(z_0\in\mathbb{R}^d\).

\begin{theorem}[Transformer simulation of closed-form diffusion]
\label{thm:cfd-realization}
There exists a choice of transformer parameters \(\theta^\star_{\mathrm{CFD}}\) such that, for every empirical dataset \(\{x_i\}_{i=1}^N\subset\mathbb{R}^d\) and every initial state \(z_0\in\mathbb{R}^d\), if \(z_L\) is generated by the closed-form diffusion recursion, then
\[
\Pi_{\mathrm{state}}
\left(
\mathcal{T}_{\theta^\star_{\mathrm{CFD}}}
\bigl(
\mathrm{Enc}(\{x_i\}_{i=1}^N,z_0)
\bigr)
\right)
=
z_L,
\]
where $L$ is the number of attention blocks. 
Moreover, fix \(\sigma\ge 0\) and \(\mathcal{E}=\{\epsilon_m\}_{m=1}^M\). There exists another choice of transformer parameters \(\theta^\star_{\sigma\text{-}\mathrm{CFD}}\), using the same prompt encoding and state readout, such that, for every empirical dataset and initial state, if \(z_L\) is generated by the smoothed recursion, then
\[
\Pi_{\mathrm{state}}
\left(
\mathcal{T}_{\theta^\star_{\sigma\text{-}\mathrm{CFD}}}
\bigl(
\mathrm{Enc}(\{x_i\}_{i=1}^N,z_0)
\bigr)
\right)
=
z_L .
\]
The parameter choices may depend on \(N,d,\tau\), and, in the smoothed case, on \(\sigma\) and \(\mathcal{E}\), but they do not depend on the realized dataset \(\{x_i\}_{i=1}^N\) or on \(z_0\).
\end{theorem}
The theorem states that the standard transformer architecture can be parameterized to execute the closed-form diffusion update rule in context. Notably, the results hold for all possible inputs while the parameters are frozen, showing that the transformer is capable of out-of-distribution generalization. We observe such generalization in Figure~\ref{fig:in-context-sampling}, where the transformer generates samples from a distribution distinct from the training data. In addition to this out-of-distribution generation, the theorem establishes uniform in-context generalization over the data values: for any fixed context length \(N\), dimension \(d\), and sampling schedule, the same transformer parameters work for every realized dataset \(\{x_i\}_{i=1}^N\) and every initial state \(z_0\).

The proof of Theorem~\ref{thm:cfd-realization} builds on the well-established computational capabilities of softmax attention layers~\cite{vaswani2017attention}, demonstrating that they can implement the iterative closed-form diffusion method. Prior work on in-context learning has similarly shown that transformers can implement various iterative algorithms, including gradient descent for regression~\cite{von2023transformers,ahn2023transformers} and temporal difference learning for reinforcement learning~\cite{wang2024transformers}. However, these studies rely on linear attention layers to simplify the analysis.
What distinguishes our work from these prior studies is the use of standard softmax attention in place of linear attention~\cite{ahn2023transformers}. In fact, the softmax normalization is essential to our construction: the closed-form diffusion update requires responsibility weights, which are precisely the normalized weights produced by softmax attention. Such normalization is omitted in linear attention due to the challenges it poses for theoretical analysis.
\begin{remark}[Relation to Rosu et al.~\cite{rosu2026softmax}]
Rosu et al.~\cite{rosu2026softmax} show that transformers can implement
in-context denoising steps using a modified attention operator, denoted
$\operatorname{Attn}$, which incorporates an RBF-type modification. In
contrast, our construction uses the standard softmax self-attention
mechanism employed by transformers in practice. The norm-dependent
quantities required to compute the Gaussian-mixture responsibilities are
included explicitly in our input encoding, rather than recovered through
a modification of the attention mechanism. Thus, our closed-form
diffusion result should be viewed as a warm-up showing that the original
transformer architecture, without the RBF modification of
\cite{rosu2026softmax}, can also implement closed-form diffusion models
from in-context samples.

Together, these results show that transformers are expressive enough to
implement several sampling algorithms. Expressivity alone, however, does
not determine which algorithm best explains the computations learned by
a trained transformer. Our main objective is to identify a generative
mechanism that captures the layer-wise evolution of representations in
pretrained language models. Closed-form diffusion does not reproduce the
observed U-shaped dynamics, in which the empirical embedding distribution
first approaches a uniform distribution and subsequently becomes
non-uniform. Moreover, the unsmoothed empirical closed-form diffusion
model targets the empirical distribution and can therefore regenerate
in-context samples. These observations motivate our analysis of
Estimation-Free Sampling (EFS), whose forward and backward dynamics motivate a qualitative analogy with the observed geometry. The original EFS work establishes population sampling results under its mean-field and limiting assumptions~\cite{daneshmand2025data}. Those results are distinct from our approximation theorem for a fixed finite, explicit discretization.
\end{remark}
\section{Mechanistic Analysis of Pretrained Transformers}
\label{sec:mechanistic-analysis}

We established the expressivity of transformers to implement closed-form diffusion models. In this section, we investigate the internal mechanism of sampling in pretrained transformers.  Our goal is not to claim that pretrained language models exactly implement a sampling algorithm. Instead, we study how they shape word embedding distribution for sampling. 

In-context sampling can be formulated as generating new words from a semantic topic, such as animals, cities, or foods, given random words from the topic. We construct prompts by sampling distinct words without replacement from a common semantic category, as detailed in Appendix~\ref{app:uniformization_models_results}. The words within a prompt are therefore dependent, and contextualization introduces further dependence among their hidden states. Then, we analyze how the distribution of word embeddings changes across the layers. Specifically, we normalize the embeddings and measure their distance to the uniform distribution on the sphere. We use squared maximum mean discrepancy, $\mathrm{MMD}^2$, as the distance metric; MMD is a standard kernel-based discrepancy between probability distributions~\cite{gretton2012kernel}. We report the off-diagonal Gaussian-kernel \(\mathrm{MMD}^2\) statistic as a descriptive finite-cloud comparison. Because contextual hidden states are dependent, we do not claim that it is an unbiased estimator of a population discrepancy. Lower values indicate a smaller measured discrepancy to the chosen reference; they do not establish convergence to a uniform distribution. 

\begin{figure}[h!]
    \centering
\begin{tabular}{c c c}
     \includegraphics[width=0.25\linewidth]{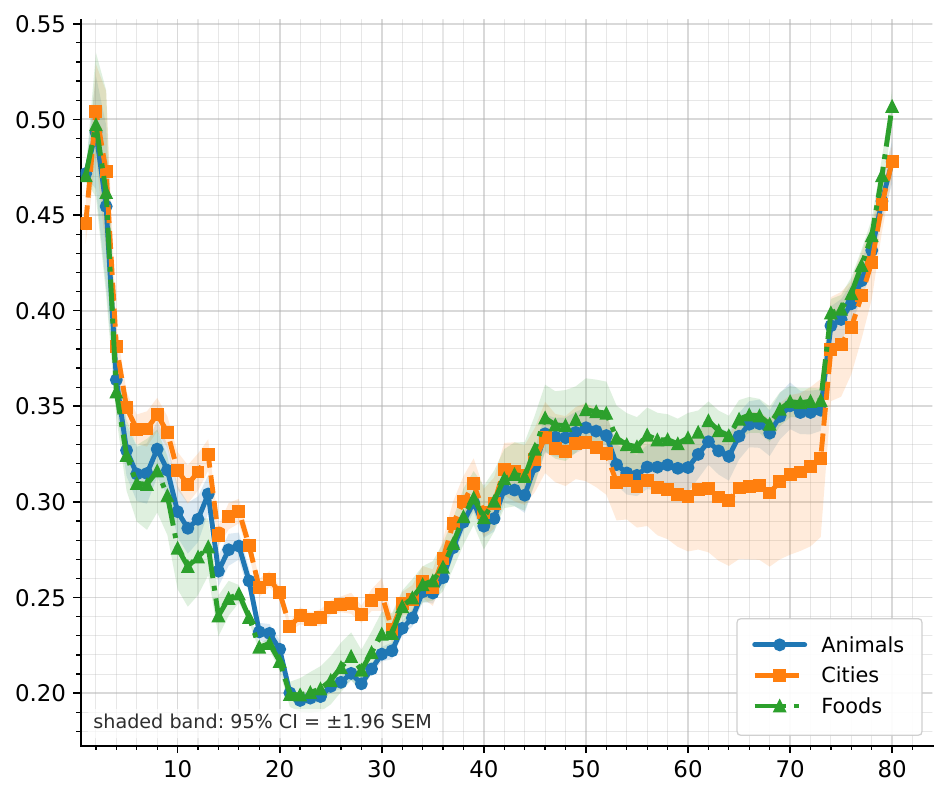} &  \includegraphics[width=0.25\linewidth]{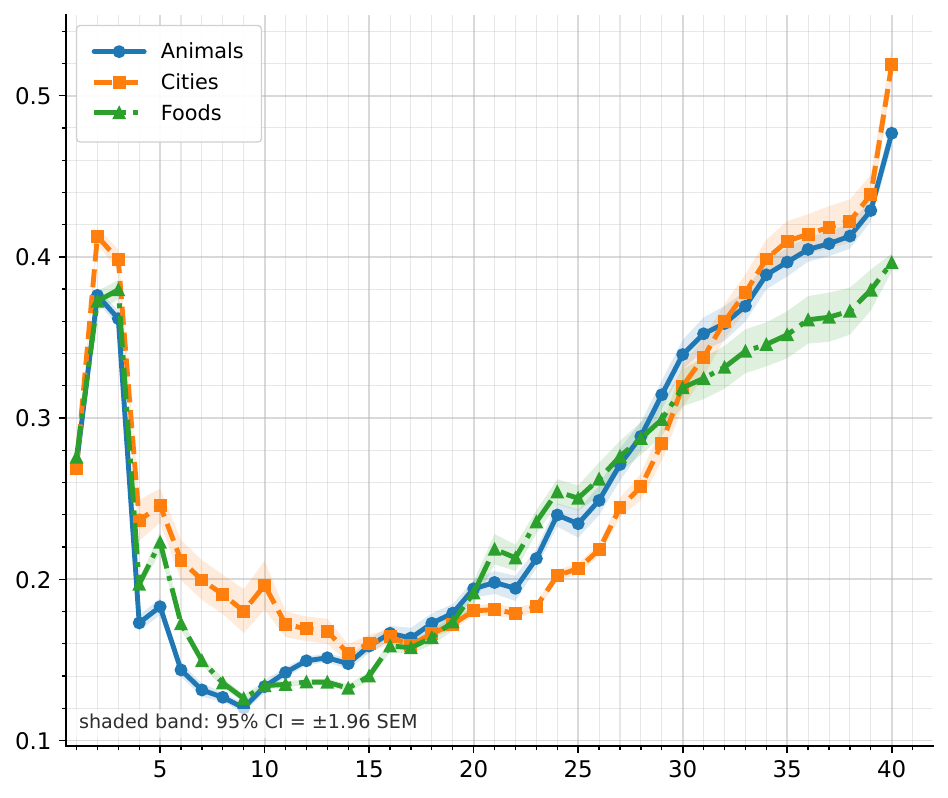} & \includegraphics[width=0.25\linewidth]{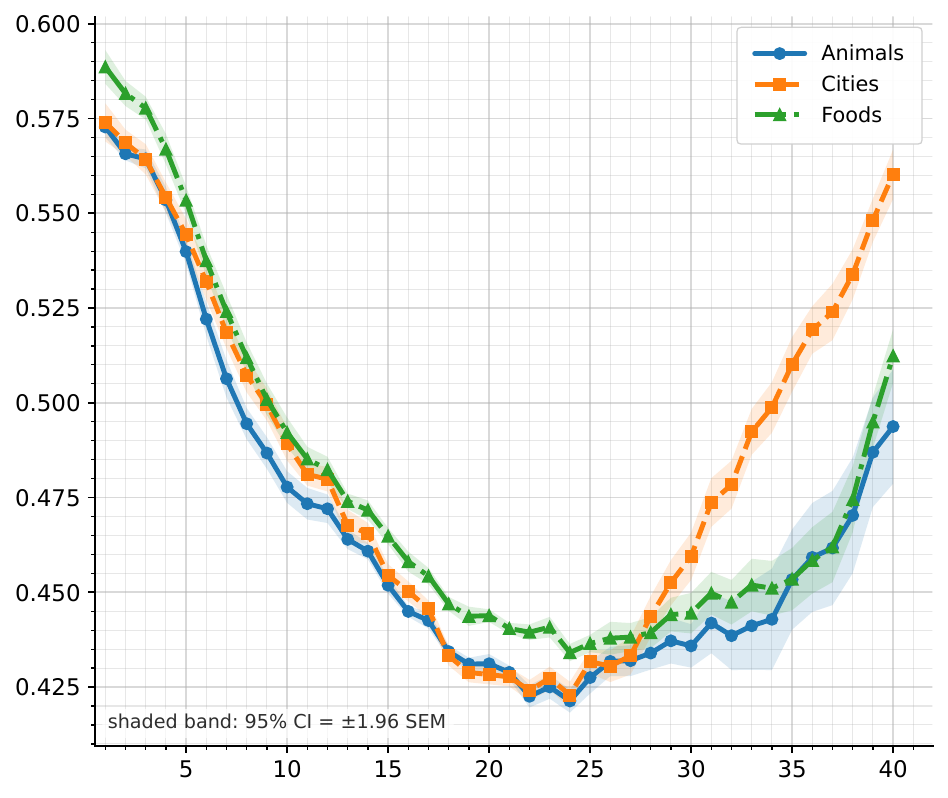}\\
    Llama-3.3-70B-Instruct~\cite{grattafiori2024llama}  & Openlm-Llama-13b~\cite{openlm2023openllama} & Cerebras-GPT-13B~\cite{dey2023cerebras} 
\end{tabular}
    \caption{\footnotesize{\textbf{Mechanism of sampling in pretrained language models.} y-axis:
    Layer-wise \(\mathrm{MMD}^2\) distance between normalized token embeddings and the uniform distribution on the sphere. $x$-axis: the layer index of word embeddings. The models are prompted with distinct words sampled without replacement from categories animal (blue), city (orange), and food (green).  Several models exhibit a U-shaped profile: intermediate layers move closer to the uniform reference distribution.}
    }
    \label{fig:mmd-uniform-iid}
\end{figure}
\begin{figure}[h!]
    \centering
\includegraphics[width=1.0\linewidth]{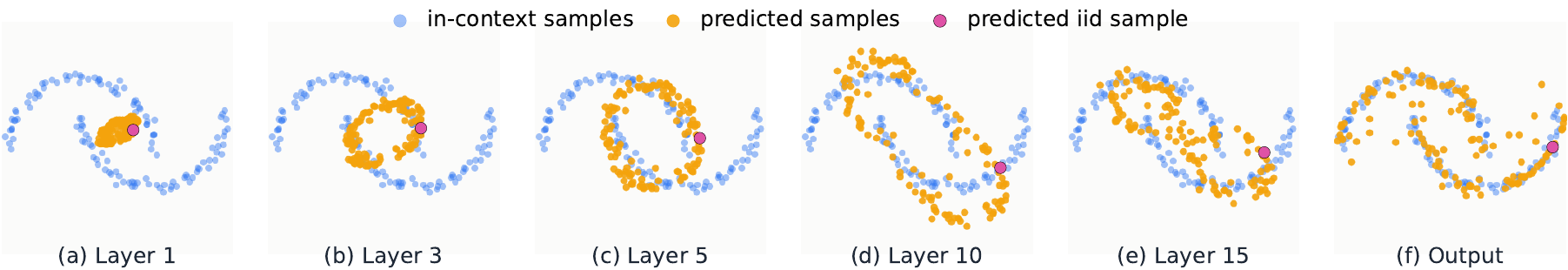}
  \caption{\footnotesize{\textbf{Uniform bias for in-context sampling.} A small GPT-2-style model is trained to reproduce samples from the two-moons 
dataset (see Appendix~\ref{app:experimental-details} for details). Orange points 
represent intermediate embeddings extracted at layers 1, 3, 5, 10, 15, and output layers, illustrating the progressive data evolution across 
transformer layers at test time. } }
    \label{fig:geometry-evolution}
\end{figure}

 We observe a two-stage pattern for three pretrained foundation models: the measured discrepancy to the uniform spherical reference decreases in intermediate layers and increases near the output. This observation holds for two variants of LLaMA~\cite{grattafiori2024llama, openlm2023openllama} and Cerebras-GPT~\cite{dey2023cerebras} in Figure~\ref{fig:mmd-uniform-iid}. To illustrate this dynamic, we present a scatter plot of word embeddings for a small trained model across the layers in Figure~\ref{fig:geometry-evolution}, where each point represents one word. The middle-layer clouds visibly spread out; this visualization does not establish convergence to a uniform distribution. 

\begin{figure}[h!]
\centering
\begin{tabular}{c c c}
\includegraphics[width=0.25\linewidth]{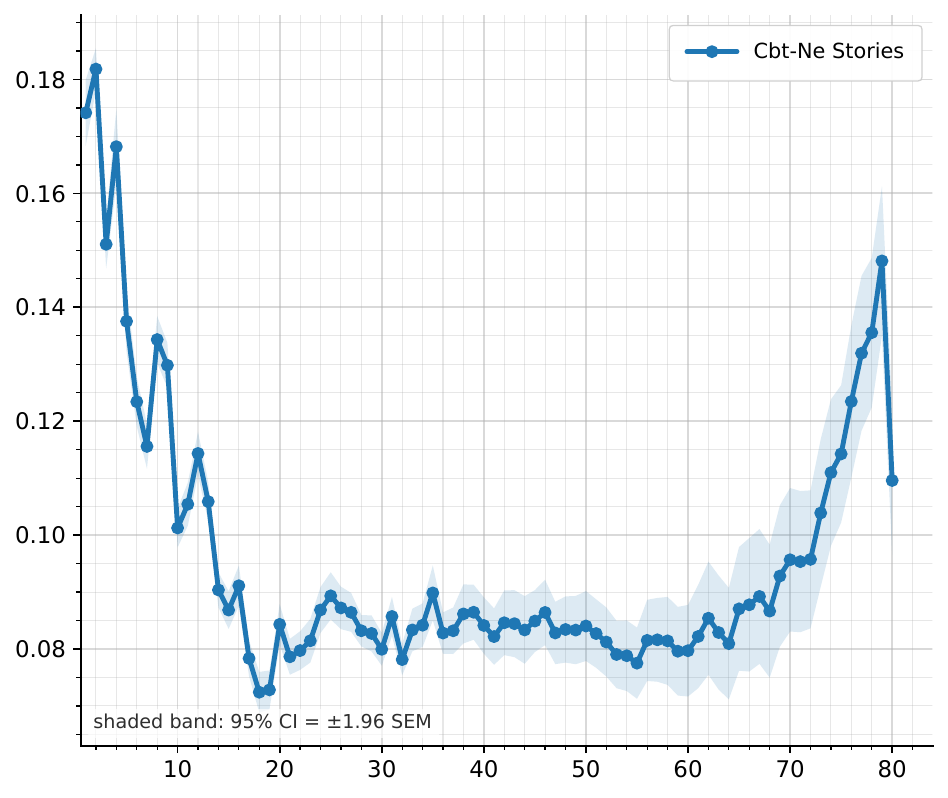} &  \includegraphics[width=0.25\linewidth]{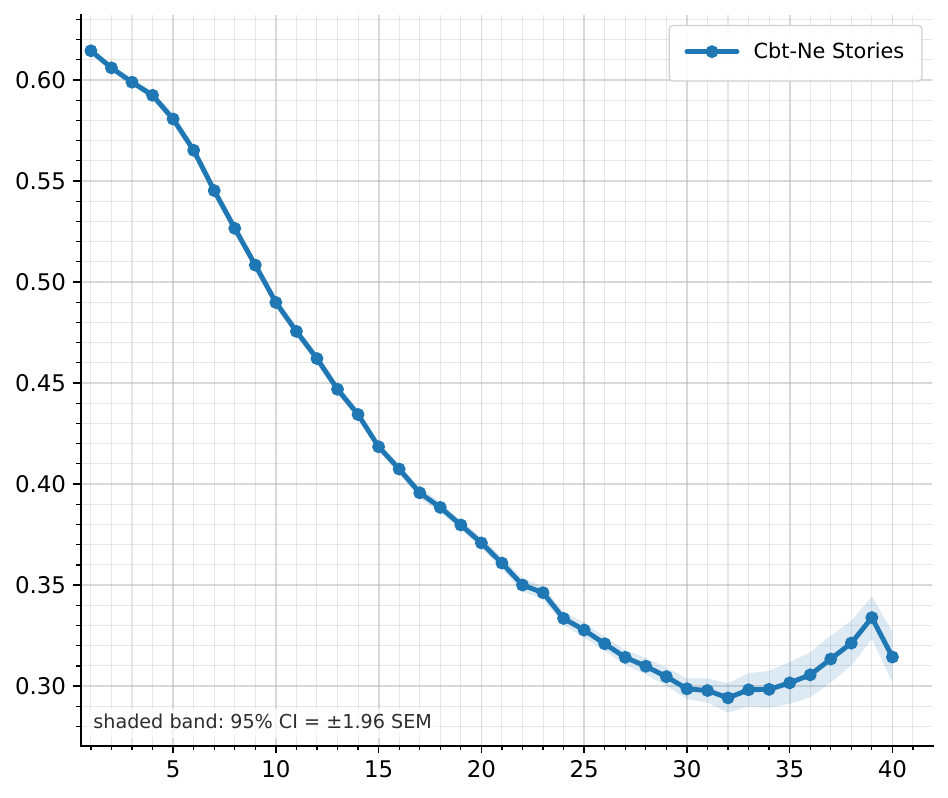} & \includegraphics[width=0.25\linewidth]{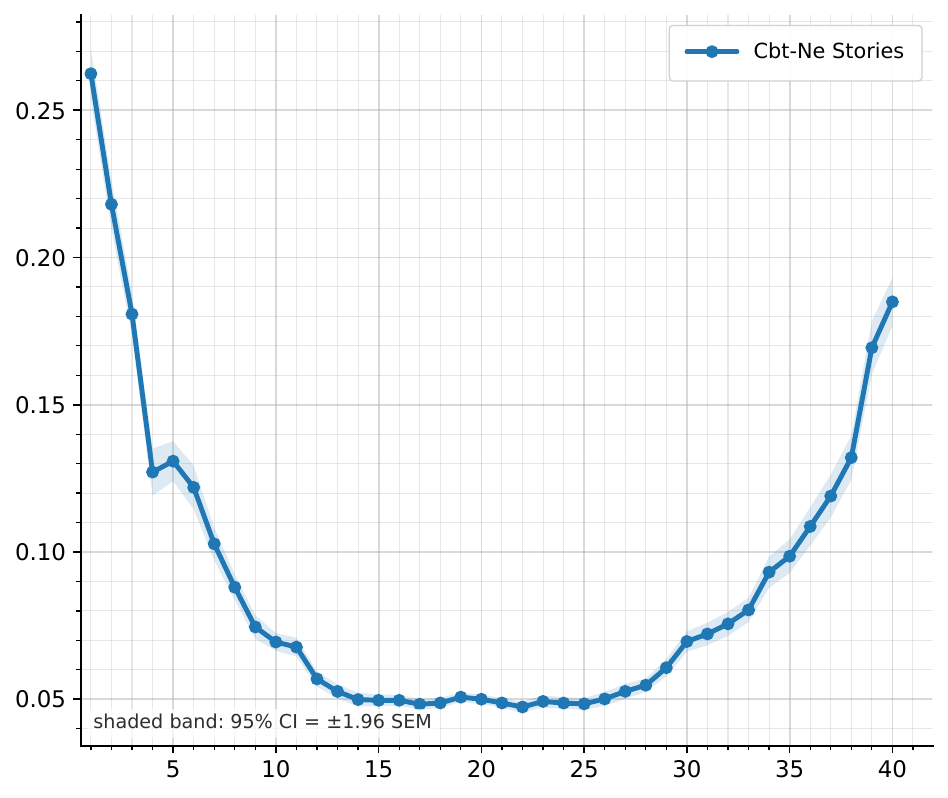}\\
   Llama-3.3-70B-Instruct~\cite{grattafiori2024llama} & Cerebras-GPT-13B~\cite{dey2023cerebras} & Openlm-Llama-13b~\cite{openlm2023openllama}
\end{tabular}

    \caption{\footnotesize{\textbf{Uniform intermediate representations in pretrained 
language models.} \textit{y}-axis: layerwise $\mathrm{MMD}^2$ between 
normalized token embeddings and the uniform distribution on the unit sphere. 
\textit{x}-axis: transformer layer index. Models are prompted with sentences 
from the CBT dataset, where each repeated word is represented by a single 
embedding vector.}}
    \label{fig:mmd-cbt-stories}
\end{figure}

We observe a similar pattern when pretrained language models are 
prompted with natural sentences rather than distinct semantic-category words.
Specifically, we prompt a pretrained model with natural sentences 
from the CBT dataset~\cite{hill2015goldilocks}.
Figure~\ref{fig:mmd-cbt-stories} plots the MMD distance between word 
embeddings and samples drawn uniformly from the unit sphere.
We use only one word embedding for repeated words when  
computing the MMD distance. 

\section{Physics of In-context Sampling} 

\subsection{Observation: Embeddings as Interacting Particles}
While proving the sampling capability, Theorem~\ref{thm:cfd-realization} cannot explain the U-shaped mechanism of sampling in pretrained models discussed in the last section. This is due to the fact that closed-form diffusion does not change the empirical data distribution. We use tools from interacting particles to provide insights into the shaping of the embedding distribution for sampling. Consider the following interacting energy defined over word embeddings
\begin{equation}
E(x_1,\ldots,x_N)
=
\frac{1}{N(N-1)}
\sum_{i=1}^N
\sum_{\substack{j=1\\ j\ne i}}^N
W_{\epsilon}^{(s)}(x_i-x_j).
\end{equation}
where \(N\ge2\), \(x_1,\dots,x_N\in\mathbb{R}^{d}\), \(\epsilon>0\), and \(s>-2\). We also write \(E_{N,\epsilon}=E\), with \(s\) fixed. For \(s\ne0\), the interaction potential is 
\[
W_{\epsilon}^{(s)}(r)
=
\frac{1}{2}\|r\|^2
+
\frac{1}{s(\|r\|^2+\epsilon)^{s/2}},
\qquad r\in\mathbb{R}^d.
\]
The logarithmic case requires subtracting an additive constant before taking the limit:
\[
W_{\epsilon}^{(0)}(r)
=\lim_{s\to0}\left(W_{\epsilon}^{(s)}(r)-\frac1s\right)
=\frac12\|r\|^2-\frac12\log(\|r\|^2+\epsilon).
\]
This renormalization leaves the interaction force unchanged. Results for continuum minimizers depend on the dimension and potential: for the unregularized logarithmic interaction with quadratic attraction, a uniform measure on a suitable sphere is a minimizer in dimensions \(d\ge4\)~\cite{frank2025minimizers}. Such results do not imply convergence of finite-particle gradient descent with \(\epsilon>0\) to a uniform distribution. We use the regularized energy as a diagnostic of the embedding geometry.

\begin{figure}[h!]
    \centering
    \begin{tabular}{c c c}
        \includegraphics[width=0.25\linewidth]{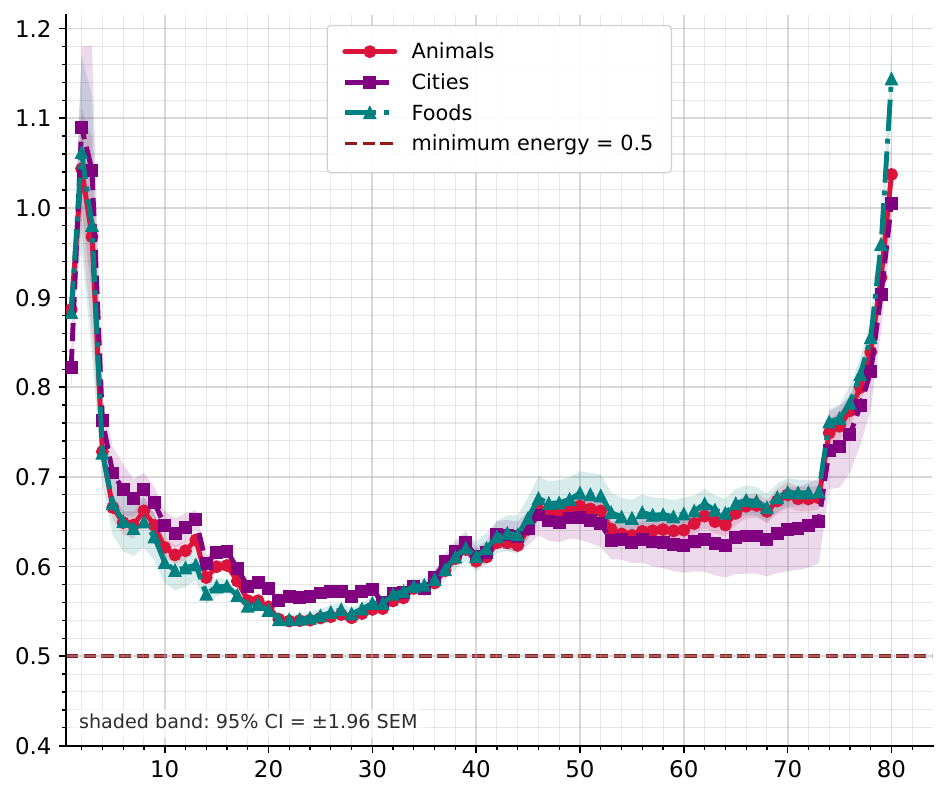}  & \includegraphics[width=0.25\linewidth]{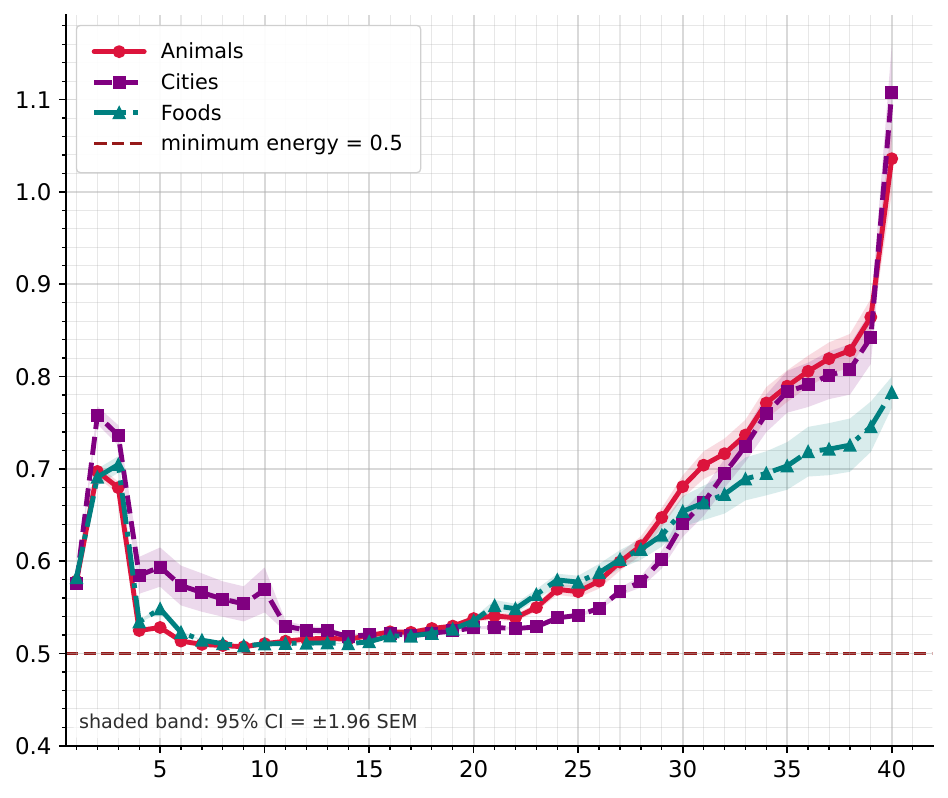} &  \includegraphics[width=0.25\linewidth]{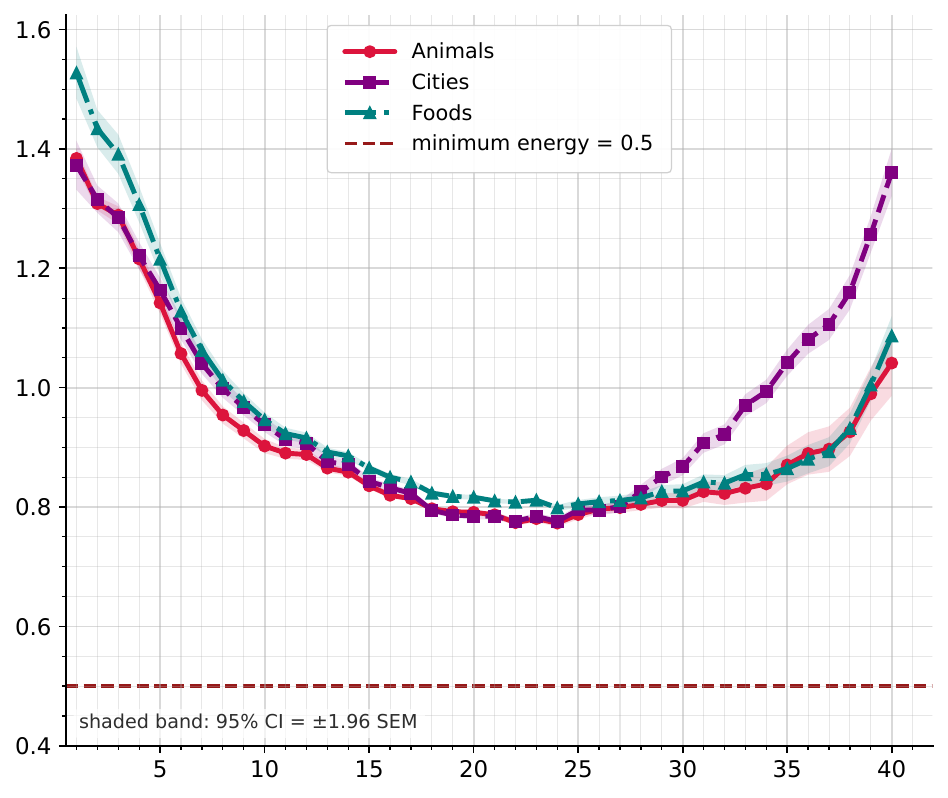}\\
        Llama-3.3-70B-Instruct~\cite{grattafiori2024llama} & Openlm-Llama-13b~\cite{openlm2023openllama} & Cerebras-GPT-13B~\cite{dey2023cerebras}
    \end{tabular}

    \caption{\footnotesize{\textbf{Repulsive-interactive energy of word embeddings.} $y$-axis: the energy function $E$ for word embeddings across the layers of a pretrained transformer. $x$-axis: the layer index. We use the renormalized logarithmic potential \(W_{\epsilon}^{(0)}\) and normalize the word embeddings to norm \(0.78\). The red dashed line at \(0.5\) is a reference level from the minimum of the unregularized logarithmic pair potential; it is not a general minimum of the regularized finite-particle energy.  }}
    \label{fig:energy-function-iid}
\end{figure}

We compute the energy of word embeddings where $x_1, \dots, x_N$ are word embeddings across layers of pretrained transformers. When the input prompt consists of distinct words from the same topic, Figure~\ref{fig:energy-function-iid} shows that the energy function produces a U-shaped plot across the layers, similar to the MMD distance plot in Figure~\ref{fig:mmd-uniform-iid}. Energy optimization also motivates EFS~\cite{daneshmand2025data}. We show that a transformer can approximate a finite explicit discretization of the corresponding forward--backward computation, providing a computational analogy for these empirical observations.  
\subsection{Theory: Energy-based sampling with particle gradient descent}
\label{sec}

Estimation-Free Sampling (EFS)~\cite{daneshmand2025data} combines forward particle optimization with backward transport. We analyze the following finite, explicit forward--backward discretization inspired by EFS. The backward step below is reverse-time Euler, not the exact inverse of a gradient-descent step; the original EFS algorithm instead computes inverse steps through an inner optimization.  
For fixed \(N\ge2\), \(d,K\ge1\), \(\gamma>0\), \(s>-2\), and \(\epsilon>0\), the forward step applies gradient descent: 
\[
x_i^{(j+1)}
=
x_i^{(j)}
-
\gamma
\nabla_{x_i}E_{N,\epsilon}
\bigl(x_1^{(j)},\ldots,x_N^{(j)}\bigr),
\qquad
j=0,\ldots,K-1.
\]
Energy decreases under the usual step-size condition \(0<\gamma<2/L_E\), where \(L_E\) bounds the Lipschitz constant of the energy gradient on a domain containing the update segments. Neither this condition nor compactness alone guarantees a uniform terminal cloud; for example, coincident particles are stationary for \(\epsilon>0\). Given a reference point \(z^{(K)}\), define the tagged-point potential
\[
V_{N,\epsilon}(z;x_1,\ldots,x_N)
=c_N\sum_{i=1}^N W_{\epsilon}^{(s)}(z-x_i),
\qquad c_N=\frac{2}{N(N-1)}.
\]
This normalization gives \(\nabla_z V_{N,\epsilon}(x_i;x_1,\ldots,x_N)=\nabla_{x_i}E_{N,\epsilon}(x_1,\ldots,x_N)\). 
The backward phase evolves the reference point through the explicit recurrence, for \(j=K,\ldots,1\),
\[
z^{(j-1)}
=
z^{(j)}
+
\gamma
\nabla_z V_{N,\epsilon}
\bigl(z^{(j)};x_1^{(j)},\ldots,x_N^{(j)}\bigr).
\] The final generated sample is \(z^{(0)}\). We collect the fixed EFS parameters in \(\tau=(K,\gamma,s,\epsilon)\), together with any fixed schedule choices.

This two-stage computation motivates an analogy with the observed layerwise geometry. The next theorem establishes uniform approximation of the finite recurrence, without asserting exact inversion, recovery of the population distribution, or a U-shaped energy trajectory.

\begin{theorem}[Transformer simulation of an explicit EFS discretization]
\label{thm:efs-realization}
Fix the parameters above and a compact set \(\mathcal{K}\subset\mathbb{R}^d\) containing all particle and reference iterates, including the initial values, for every admissible input. Then, for every \(\varepsilon>0\), there exist a finite transformer depth, finite feedforward hidden widths, and parameters \(\theta^\star_{\mathrm{EFS}}\) such that, for every admissible input \(x_1^{(0)},\ldots,x_N^{(0)},z^{(K)}\), if \(z^{(0)}\) is generated by the explicit forward--backward recursion above, then
\[
\left\|
\Pi_{\mathrm{state}}
\left(
\mathcal{T}_{\theta^\star_{\mathrm{EFS}}}
\bigl(
\mathrm{Enc}_{\mathrm{EFS}}(x_1^{(0)},\ldots,x_N^{(0)},z^{(K)})
\bigr)
\right)
-
z^{(0)}
\right\|
\leq \varepsilon .
\]
The depth, feedforward widths, and parameter choice \(\theta^\star_{\mathrm{EFS}}\) may depend on \(N,d,K,\tau,\varepsilon\), and the compact domain \(\mathcal{K}\), but not on the realized particle cloud \(x_1^{(0)},\ldots,x_N^{(0)}\) or reference point \(z^{(K)}\).
\end{theorem}

The theorem shows that a residual softmax transformer can approximate this finite particle-based computation in context using the encoding in Appendix~\ref{subsec:efs-encoding}. It controls the terminal state readout. It does not prescribe the interaction energy of the full hidden-state vectors across layers, which also contain trajectory memory and scratch coordinates; the U-shaped curves in Figure~\ref{fig:energy-function-iid} remain empirical observations.   

\section{Experiments}
\label{sec:experiments}

The theory above shows that transformer depth can implement iterative sampling algorithms in context. We now test whether this perspective is visible in trained and pretrained transformers. Our experiments are designed around two questions: (i) can a transformer generate samples from a distribution specified only by the prompt, and (ii) do hidden states exhibit an intermediate transport-like phase consistent with the EFS picture?

\paragraph{In-context sampling in a controlled setting.}
We train a small GPT-2-style decoder-only transformer~\cite{radford2019language} on two-dimensional i.i.d.\ point-cloud sequences using a causal next-token objective. Figure~\ref{fig:in-context-sampling} shows a held-out compositional test: the training distribution contains the face boundary and eyes, but no smile component. When smile-shaped points are supplied in context, the model generates new samples along the same crescent geometry. This suggests that the prompt acts as an empirical distribution, rather than merely selecting among memorized training components.

\paragraph{Layerwise transport in a learned sampler.}
We train the same architecture on two-moons data and project intermediate hidden states back to the data space. The resulting geometry evolves in a transport-like manner: early layers concentrate the generated points, middle layers spread them toward a more uniform configuration, and later layers recover the structured two-moons geometry specified by the context. A full layer-by-layer visualization is provided in Appendix~\ref{app:layerwise_geometry}.

\paragraph{Uniformization in pretrained language models.}
We next examine pretrained language models using prompts of distinct semantic-category words sampled without replacement, such as animals, foods, and cities. At each layer, we normalize token embeddings and measure their off-diagonal RBF \(\mathrm{MMD}^2\) statistic relative to the uniform distribution on the sphere. Figure~\ref{fig:mmd-uniform-iid} shows a U-shaped profile across several models: intermediate layers move closer to the uniform reference, while later layers move away from it toward structured, topic-dependent representations. The same qualitative behavior appears for natural text prompts from the CBT dataset~\cite{hill2015goldilocks} in Figure~\ref{fig:mmd-cbt-stories}. Additional results and experimental details are deferred to Appendix~\ref{app:uniformization_models_results}.

\paragraph{Energy-based evidence.}
Finally, we evaluate the EFS-style interaction energy on the same hidden-state clouds. Figure~\ref{fig:energy-function-iid} shows that the energy follows the same qualitative middle-layer regularization pattern as the MMD curves. This supports the interacting-particle interpretation: intermediate layers move representations toward a lower-energy, more uniform configuration, while later layers recover structured, topic-dependent geometry. 

\section{Discussion and Limitations}
\label{sec:limitations}

\begin{figure}[h]
    \centering

    \begin{subfigure}[t]{0.19\textwidth}
        \centering
        \includegraphics[width=\linewidth]{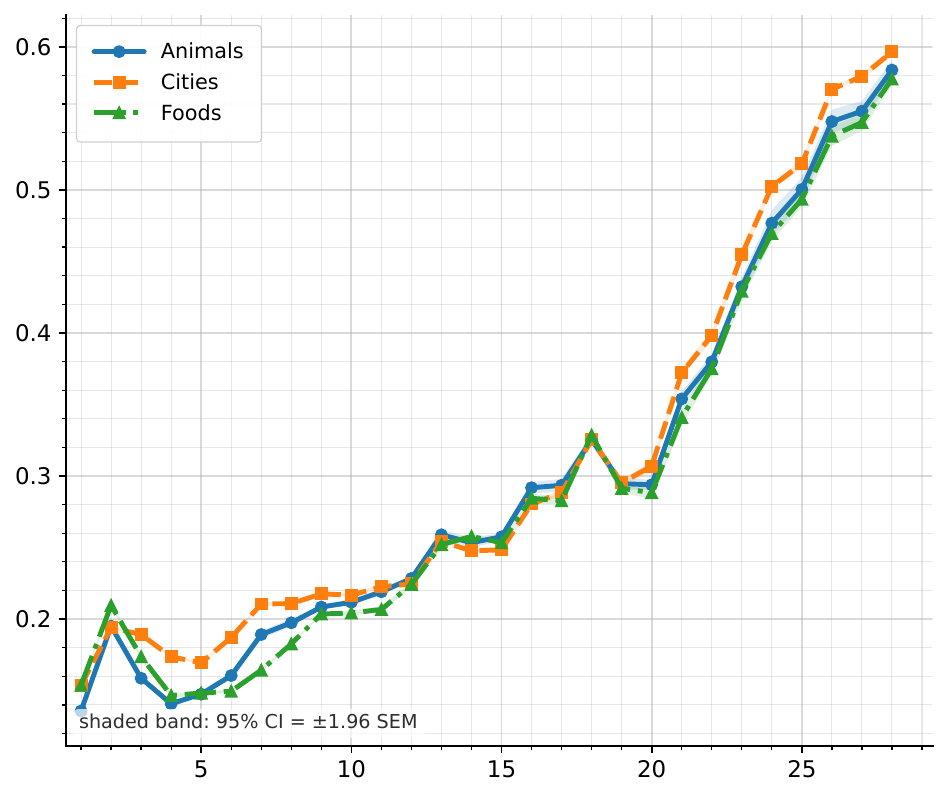}
        \caption{1.5B}
        \label{fig:qwen-family-1p5b}
    \end{subfigure}
    \hfill
    \begin{subfigure}[t]{0.19\textwidth}
        \centering
        \includegraphics[width=\linewidth]{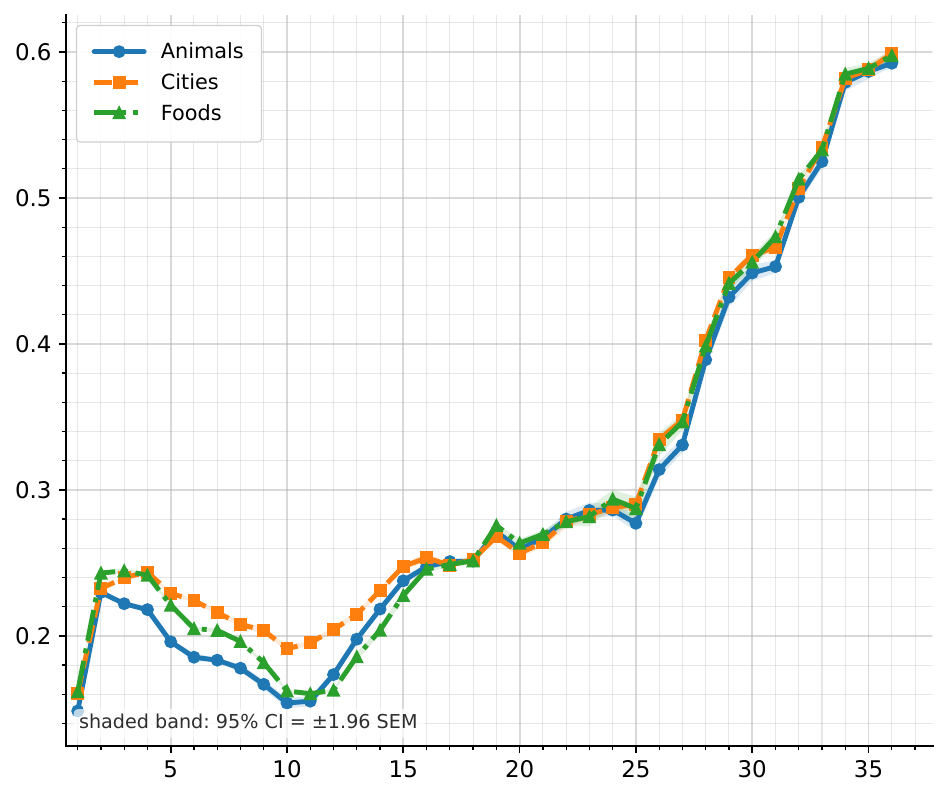}
        \caption{3B}
        \label{fig:qwen-family-3b}
    \end{subfigure}
    \hfill
    \begin{subfigure}[t]{0.19\textwidth}
        \centering
        \includegraphics[width=\linewidth]{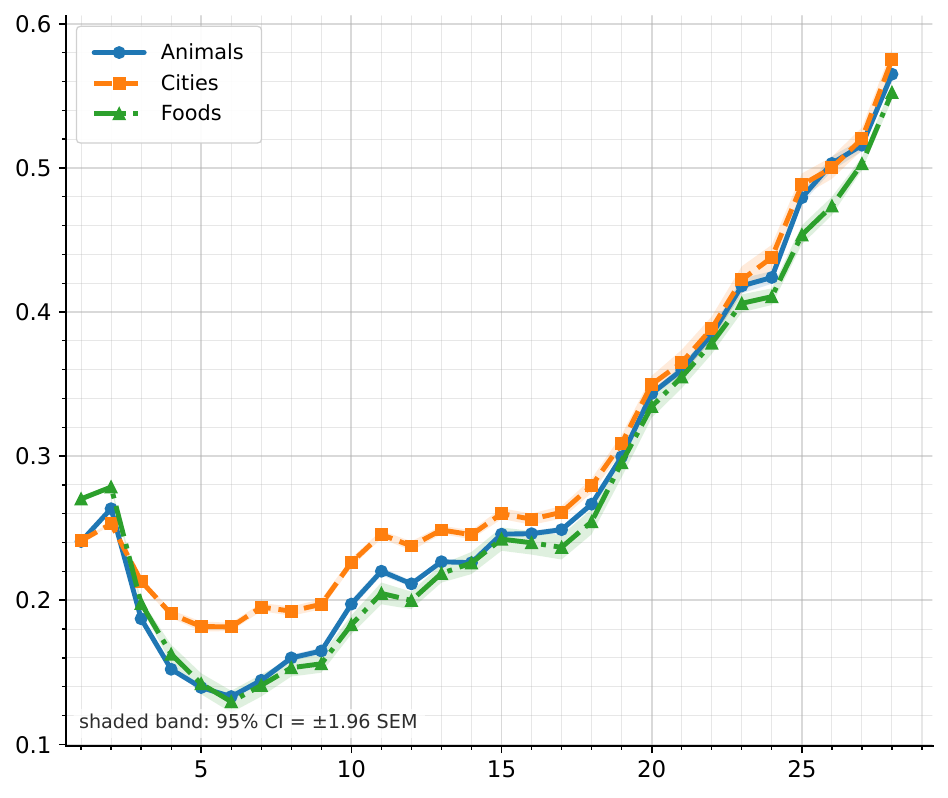}
        \caption{7B}
        \label{fig:qwen-family-7b}
    \end{subfigure}
    \hfill
    \begin{subfigure}[t]{0.19\textwidth}
        \centering
        \includegraphics[width=\linewidth]{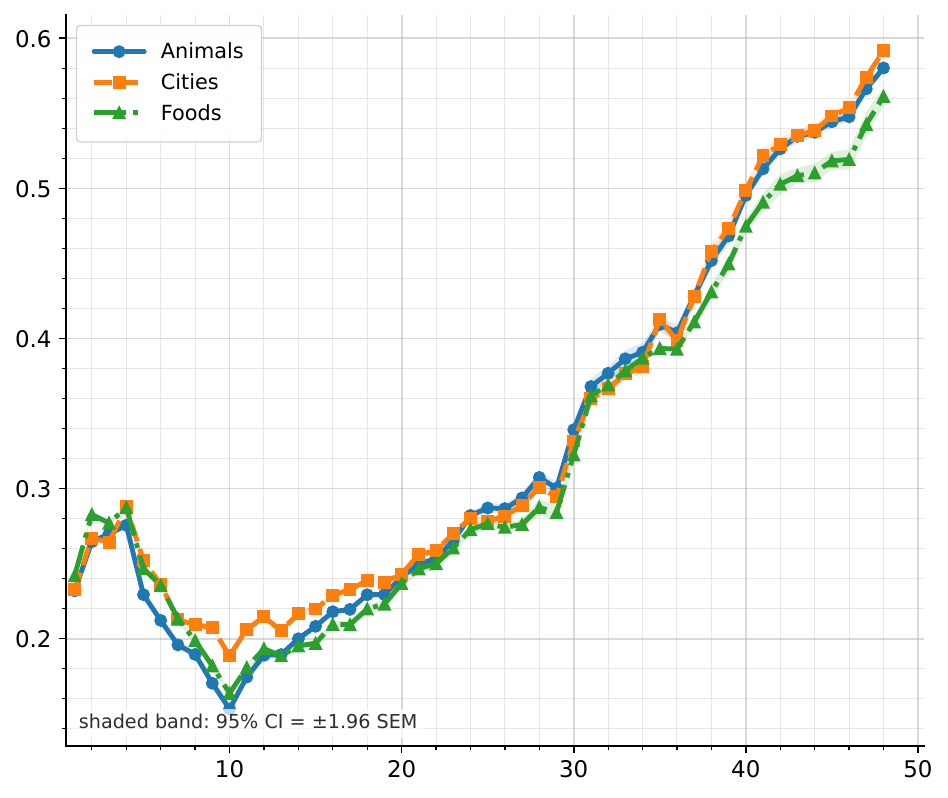}
        \caption{14B}
        \label{fig:qwen-family-14b}
    \end{subfigure}
    \hfill
    \begin{subfigure}[t]{0.19\textwidth}
        \centering
        \includegraphics[width=\linewidth]{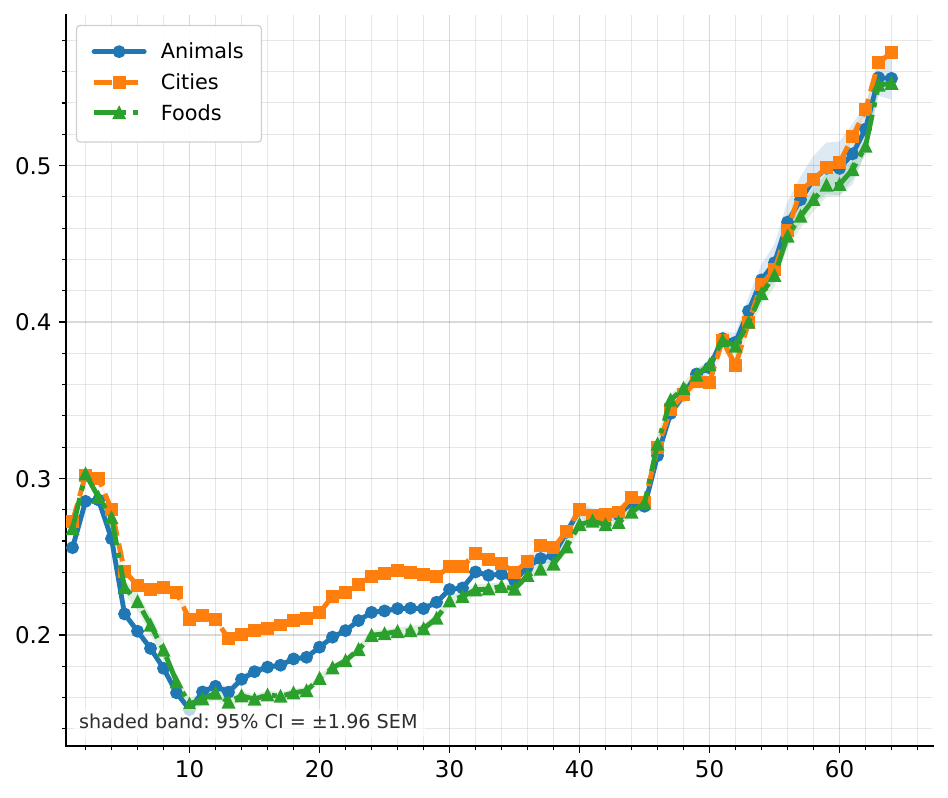}
        \caption{32B}
        \label{fig:qwen-family-32b}
    \end{subfigure}
    
    \caption{
   Qwen2.5 family ~\cite{qwen2024qwen2}  on semantic-category prompts sampled without replacement. 
Unlike the clearer U-shaped profiles observed in larger pretrained models, the Qwen2.5 family exhibits a weaker and less pronounced U-shaped pattern. 
The curves still suggest a middle-layer movement toward a uniform spherical reference distribution followed by a later movement away from it, but the effect is shorter and less robust across model scales.
    }
    \label{fig:fail}
\end{figure}

\paragraph{Failure mode across model scales.}
The uniformization effect is not equally pronounced across all models. In the Qwen2.5 family ~\cite{qwen2024qwen2}, smaller models exhibit only a short or weak movement toward the uniform reference, whereas larger variants show a clearer two-stage profile as shown in Figure~\ref{fig:fail}. This provides a concrete failure mode for the proposed mechanism: when the model has insufficient effective depth or capacity, the intermediate transport phase may be incomplete. This observation is consistent with our theoretical construction, where transformer depth controls the number of iterative computation steps available to the model. Additional failure-mode experiments for Qwen2.5 and also GPT-2 families are provided in Appendix~\ref{app:failure_mode}.

\paragraph{Gap between expressivity and mechanism.}
Our theoretical results are expressivity statements: they show that some transformer parameters can implement closed-form diffusion exactly and approximate a finite explicit EFS-inspired recursion from in-context samples. They do not prove that pretrained language models execute these algorithms internally, nor does Theorem~\ref{thm:efs-realization} guarantee a uniform intermediate cloud or population-level sampling accuracy. Our experiments provide compatible mechanistic evidence, including middle-layer movement toward a uniform reference distribution and a decrease in EFS-style interaction energy. Still, these observations do not uniquely identify the internal algorithm. This gap is common in in-context learning: prior work shows that transformers can learn regression functions~\cite{garg2022what}, implement gradient-descent-like computations~\cite{ahn2023transformers,von2023transformers}, or realize temporal-difference updates~\cite{wang2024transformers}, but such results mainly characterize representational capacity or behavior in controlled settings. Similarly, our work identifies sampling algorithms that transformers can implement and shows compatible layerwise dynamics, while leaving a full characterization of the mechanism to future work.
\paragraph{Generative AI statement.}
The authors used generative AI tools to assist with code implementation, proof development and verification, and language editing. All mathematical arguments, experimental results, and generated text were independently reviewed and validated by the authors, who take full responsibility for the content of this paper.

\newpage
\bibliography{refs}
\bibliographystyle{unsrtnat}

\appendix

\section*{Technical appendices and supplementary material}
\section{Related Work}\label{app:related}

\paragraph{In-context learning.}
In-context learning refers to the ability of a model to adapt its behavior at inference time using only information supplied in the prompt, without updating its parameters. This phenomenon was popularized by large language models~\cite{brown2020language,chowdhery2023palm,achiam2023gpt,touvron2023llama,touvron2023llama2,grattafiori2024llama} and has since become a central topic in the study of transformer-based models~\cite{dong2024survey}. Much of the empirical literature studies how performance depends on prompt construction, including the choice, ordering, and formatting of demonstrations~\cite{liu2022makes,lu2022fantastically,wu2023selfadaptive}. Complementary work studies mechanisms behind in-context learning, including induction heads, attention-based information flow, Bayesian inference, and gradient-descent-like computation inside transformers~\cite{olsson2022induction,wang2023label,xie2022explanation,zhang2023what,dai2023why,von2023transformers,ahn2023transformers,mahankali2023one}.

Most theoretical studies of in-context learning focus on prediction. In the standard formulation, the prompt contains input--output examples from an unknown task, and the model predicts the output for a new query. Garg et al.~\cite{garg2022what} showed that transformers trained from scratch can learn simple function classes from examples in the prompt, including linear functions, sparse linear functions, decision trees, and shallow neural networks. Related work studies whether transformers implement particular learning algorithms in context, including ridge regression, gradient descent, higher-order optimization, and algorithm selection~\cite{akyurek2023what,li2023transformers,bai2023transformers,fu2023transformers}. Our work follows this algorithmic view, but changes the target from prediction to generation: instead of asking whether a transformer can predict a label or function value, we ask whether it can execute a sampler in context.

\paragraph{Transformers as algorithm executors.}
A growing line of work studies transformers as models capable of executing structured computations. Prior work has shown that transformer depth and attention can implement iterative optimization procedures, temporal-difference style updates, and discrete optimal transport or sorting-type computations~\cite{ahn2023transformers,wang2024transformers,daneshmand2024provable,daneshmand2026context}. These results suggest that attention and depth can act as computational resources: attention aggregates information from the prompt, while depth implements repeated updates.

Our results extend this perspective to generative algorithms. Closed-form diffusion and Estimation-Free Sampling are not prediction rules; they are iterative procedures that transform an initial state into a sample. Showing that frozen transformers can implement these procedures in context strengthens the interpretation of transformers as general-purpose algorithm executors. The main difference from prior algorithmic ICL work is that the output of the computation is not a predicted label or regression value, but the terminal state of a sampler.

\paragraph{Diffusion and score-based generative models.}
Diffusion models generate samples by gradually transforming noise into data through an iterative reverse process~\cite{ho2020denoising}. Score-based generative modeling provides a continuous-time formulation in which a forward stochastic process maps data to noise and a reverse-time process generates data from noise using estimated score functions~\cite{song2020score}. Related samplers, including DDIM, modify the reverse process through deterministic or implicit updates~\cite{song2020denoising}. These methods have made iterative sampling a central computational paradigm in modern generative modeling.

Our work does not propose a new diffusion model or train a new score network. Instead, we isolate the computational structure of diffusion-style sampling. In the closed-form setting, the score and sampling dynamics are explicit functions of the empirical samples and the current state. This allows us to ask a representational question: can a frozen transformer implement the corresponding sampler when the empirical distribution and initial state are supplied in the prompt? Our answer is affirmative, showing that a core diffusion-style computation can be realized as in-context computation by a transformer.

\paragraph{In-context generation with diffusion models.}
There is prior work on enabling in-context learning inside diffusion models. Prompt Diffusion trains a diffusion-based model to perform vision tasks from example input--output pairs and a query image, thereby bringing in-context learning to diffusion-based generation~\cite{wang2023promptdiffusion}. The paper is the closest prior work connecting diffusion models and in-context learning, but its direction is different: it equips a diffusion model with in-context learning ability. 

Our question reverses this direction. Rather than asking whether a diffusion model can perform in-context learning, we ask whether a transformer can implement diffusion-style and particle-based samplers in context. Thus, the central object in our paper is not a diffusion architecture conditioned on examples, but a frozen transformer that executes a sampling algorithm specified by the prompt.

\paragraph{Estimation-Free Sampling.}
Estimation-Free Sampling (EFS) proposes a generative mechanism that avoids explicit score estimation and neural generative training~\cite{daneshmand2025data}. Instead of learning a score or density-dependent function, EFS uses deterministic particle optimization. Its forward step transports empirical particles toward a simple reference distribution, and its backward step maps a newly sampled reference point back toward the data distribution. The EFS paper explicitly frames this as a generation without score/function estimation, neural-network training, or noise injection in the mean-field regime. 

We include EFS because it tests whether in-context sampling is limited to diffusion-style score dynamics. Our result shows that it is not. A frozen transformer can also implement a particle-optimization sampler in context. This broadens the scope of the theory: transformers can realize not only closed-form diffusion updates, but also deterministic particle-based generative algorithms.

\paragraph{Positioning.}
Taken together, prior work shows that transformers can learn from context, diffusion models can generate through iterative sampling, and particle methods can generate without explicit function estimation. Our contribution connects these threads. We introduce in-context sampling as a framework for studying generative algorithms executed by frozen transformers. In this framework, the prompt specifies the sampling instance, attention computes interactions among samples or particles, and transformer depth performs the iterative updates. This perspective broadens in-context learning from prediction to generation and places closed-form diffusion and Estimation-Free Sampling under a common computational view.

\section{Proof of Theorem~\ref{thm:cfd-realization}}
\label{app:proof-cfd-realization}

We use the prompt encoding from Section~\ref{subsec:embedding-encoding}. Each token has a block form
\[
X_i=
\bigl[
X_i^{(x)},\,
X_i^{(r)},\,
X_i^{(z)},\,
X_i^{(k)},\,
X_i^{(1)}
\bigr],
\]
where \(X_i^{(x)},X_i^{(z)},X_i^{(k)}\in\mathbb{R}^d\), and
\(X_i^{(r)},X_i^{(1)}\in\mathbb{R}\). Initially, for all tokens \(i\le N\),
\[
X_i^{(x)}=x_i,\qquad
X_i^{(r)}=\|x_i\|^2,\qquad
X_i^{(k)}=0_d,\qquad
X_i^{(1)}=1.
\]
The final data token \(i=N\) is also the state token; initially \(X_N^{(z)}=z_0\), while \(X_i^{(z)}=0_d\) for \(i<N\).

Let \(L\) be the number of transformer layers. We take \(L=S\), one layer per Euler step, and write the schedule as
\[
\tau=\{(t_\ell,h_\ell)\}_{\ell=0}^{L-1}.
\]
We prove the result by induction over layers. The invariant is that after \(\ell\) blocks, the state token stores \(z_\ell\) in its \(z\)-block, while every token retains \(x_i\), \(\|x_i\|^2\), and \(1\) in its \(x\)-, \(r\)-, and constant blocks and has a zero \(k\)-block. The other tokens' \(z\)-blocks are unrestricted. The base case follows from the prompt encoding. It remains to construct one block \(B_{t,h}\) that maps
\[
z \mapsto z+h v_t(z)
\]
and resets the scratch block.

\subsection{Attention layer convention}

A single-head attention layer is specified by fixed matrices
\[
W_Q,W_K\in\mathbb{R}^{p\times d_q},
\qquad
W_V\in\mathbb{R}^{p\times p}.
\]
For \(X\in\mathbb{R}^{N\times p}\), define
\[
q_i=X_iW_Q,\qquad
k_i=X_iW_K,\qquad
v_i=X_iW_V .
\]
Under ordinary full or causal self-attention, the final token attends to all \(N\) tokens, including itself. Hence
\[
a_{N,i}(X)
=
\frac{
\exp\!\left(\langle q_{N},k_i\rangle\right)
}{
\sum_{j=1}^N
\exp\!\left(\langle q_{N},k_j\rangle\right)
},
\qquad i=1,\ldots,N,
\]
and
\[
\mathrm{Attn}(X)_{N}
=
\sum_{i=1}^N a_{N,i}(X)v_i .
\]
The usual \(1/\sqrt{d_q}\) scaling can be absorbed into \(W_Q\) or \(W_K\), so we omit it.

\subsection{One closed-form diffusion step}

Fix \(t\in(0,1)\) and \(h>0\). By~\eqref{eq:cfd-velocity},
\[
v_t(z)
=
-\frac{1}{1-t}z
+
\frac{1}{t(1-t)}k_t(z).
\]
Thus it suffices for attention to compute \(k_t(z)\).

\paragraph{Query, key, and value projection weights.}
Let
\[
a_t=\frac{t}{(1-t)^2},
\qquad
b_t=-\frac{t^2}{2(1-t)^2}.
\]
Choose \(d_q=d+1\). With block ordering \((x,r,z,k,1)\), define
\[
W_Q^{(t)}
=
\begin{bmatrix}
0_{d\times d} & 0_{d\times 1}\\
0_{1\times d} & 0\\
a_t I_d       & 0_{d\times 1}\\
0_{d\times d} & 0_{d\times 1}\\
0_{1\times d} & 1
\end{bmatrix},
\qquad
W_K^{(t)}
=
\begin{bmatrix}
I_d           & 0_{d\times 1}\\
0_{1\times d} & b_t\\
0_{d\times d} & 0_{d\times 1}\\
0_{d\times d} & 0_{d\times 1}\\
0_{1\times d} & 0
\end{bmatrix}.
\]
Therefore
\[
q_i=X_iW_Q^{(t)}
=
\bigl[
a_tX_i^{(z)},\,X_i^{(1)}
\bigr],
\qquad
k_i=X_iW_K^{(t)}
=
\bigl[
X_i^{(x)},\,b_tX_i^{(r)}
\bigr].
\]
For the state token \(N\) and any data token \(i\le N\), including \(i=N\),
\[
\langle q_{N},k_i\rangle
=
\frac{t}{(1-t)^2}\langle z,x_i\rangle
-
\frac{t^2}{2(1-t)^2}\|x_i\|^2 .
\]
Moreover,
\[
-\frac{\|z-tx_i\|^2}{2(1-t)^2}
=
\frac{t}{(1-t)^2}\langle z,x_i\rangle
-
\frac{t^2}{2(1-t)^2}\|x_i\|^2
-
\frac{\|z\|^2}{2(1-t)^2}.
\]
The last term is independent of \(i\), so it cancels in the softmax over data tokens. Hence
\[
a_{N,i}(X)=w_i(t,z),
\qquad i=1,\ldots,N,
\]
where \(w_i(t,z)\) are the responsibility weights defined in~\eqref{eq:resp}.

The value projection is the block matrix
\[
W_V^{(t)}
=
\begin{bmatrix}
0_{d\times d} & 0_{d\times 1} & 0_{d\times d} & tI_d & 0_{d\times 1}\\
0_{1\times d} & 0 & 0_{1\times d} & 0_{1\times d} & 0\\
0_{d\times d} & 0_{d\times 1} & 0_{d\times d} & 0_{d\times d} & 0_{d\times 1}\\
0_{d\times d} & 0_{d\times 1} & 0_{d\times d} & 0_{d\times d} & 0_{d\times 1}\\
0_{1\times d} & 0 & 0_{1\times d} & 0_{1\times d} & 0
\end{bmatrix}.
\]
Thus, for a data token \(i\),
\[
v_i=X_iW_V^{(t)}
=
[0_d,0,0_d,t x_i,0],
\]
and the attention output at the state token satisfies
\[
\mathrm{Attn}(X)_{N}^{(k)}
=
\sum_{i=1}^N a_{N,i}(X)t x_i
=
\sum_{i=1}^N w_i(t,z)(t x_i)
=
k_t(z).
\]

\paragraph{Feedforward update.}
Let
\[
Y=X+\mathrm{Attn}(X).
\]
The feedforward layer is the row-wise linear map below, realizable exactly by a two-layer ReLU network using \(u=\operatorname{ReLU}(u)-\operatorname{ReLU}(-u)\):
\[
\mathrm{FF}_{t,h}(Y_i)=Y_iW_{\mathrm{FF}}^{(t,h)},
\]
where, with the same block ordering \((x,r,z,k,1)\),
\[
W_{\mathrm{FF}}^{(t,h)}
=
\begin{bmatrix}
0_{d\times d} & 0_{d\times 1} & 0_{d\times d} & 0_{d\times d} & 0_{d\times 1}\\
0_{1\times d} & 0 & 0_{1\times d} & 0_{1\times d} & 0\\
0_{d\times d} & 0_{d\times 1} & -\frac{h}{1-t}I_d & 0_{d\times d} & 0_{d\times 1}\\
0_{d\times d} & 0_{d\times 1} & \frac{h}{t(1-t)}I_d & -I_d & 0_{d\times 1}\\
0_{1\times d} & 0 & 0_{1\times d} & 0_{1\times d} & 0
\end{bmatrix}.
\]
Equivalently,
\[
\mathrm{FF}_{t,h}^{(z)}(Y_i)
=
-\frac{h}{1-t}Y_i^{(z)}
+
\frac{h}{t(1-t)}Y_i^{(k)},
\qquad
\mathrm{FF}_{t,h}^{(k)}(Y_i)
=
-Y_i^{(k)},
\]
and all other output blocks are zero. For the state token,
\[
Y_{N}^{(z)}=z,
\qquad
Y_{N}^{(k)}=k_t(z).
\]
Therefore the residual update gives
\[
B_{t,h}(X)_{N}^{(z)}
=
Y_{N}^{(z)}
+
\mathrm{FF}_{t,h}^{(z)}(Y_{N})
=
z-\frac{h}{1-t}z+\frac{h}{t(1-t)}k_t(z)
=
z+h v_t(z),
\]
and
\[
B_{t,h}(X)_{N}^{(k)}
=
Y_{N}^{(k)}
+
\mathrm{FF}_{t,h}^{(k)}(Y_{N})
=
0_d.
\]
Thus \(B_{t,h}\) implements one closed-form diffusion Euler step, preserves every token's \(x\)-, \(r\)-, and constant blocks, and resets every \(k\)-block. The other \(z\)-blocks may change, but do not enter the key or value projections.

\subsection{Completion of the induction}

For each layer \(\ell=0,\ldots,L-1\), choose
\[
B_\ell=B_{t_\ell,h_\ell}.
\]
The parameter choice \(\theta^\star_{\mathrm{CFD}}\) is obtained by stacking
\[
\mathcal{T}_{\theta^\star_{\mathrm{CFD}}}
=
B_{L-1}\circ\cdots\circ B_0 .
\]
The base case \(\ell=0\) follows from the prompt encoding. If the induction invariant holds at layer \(\ell\), then \(B_\ell\) updates
\[
z_\ell
\mapsto
z_\ell+h_\ell v_{t_\ell}(z_\ell)
=
z_{\ell+1},
\]
resets every \(k\)-block to zero, and preserves all \(x\)-, \(r\)-, and constant blocks. Hence the invariant holds at layer \(\ell+1\). By induction, after \(L\) layers the state token stores \(z_L\). Therefore
\[
\Pi_{\mathrm{state}}
\left(
\mathcal{T}_{\theta^\star_{\mathrm{CFD}}}
\bigl(
\mathrm{Enc}(\{x_i\}_{i=1}^N,z_0)
\bigr)
\right)
=
z_L,
\]
where \(z_L\) is generated by the closed-form diffusion recursion.

\subsection{Smoothed closed-form diffusion}

The smoothed case uses the same induction argument, with one modification: the one-step block computes \(k_{\sigma,t}(z)\) instead of \(k_t(z)\). Fix \(t\in(0,1)\), \(h>0\), smoothing level \(\sigma\ge 0\), and perturbations
\[
\epsilon_1,\ldots,\epsilon_M\in\mathbb{R}^d.
\]
Recall that
\[
k_{\sigma,t}(z)
=
\frac{1}{M}\sum_{m=1}^M k_t(z+\sigma\epsilon_m).
\]
Thus we compute \(k_t(z+\sigma\epsilon_m)\) for each \(m\) and average the results.

Use \(M\) attention heads. For head \(m\), keep \(W_K^{(t)}\) and \(W_V^{(t)}\) as above and replace \(W_Q^{(t)}\) by
\[
W_{Q,m}^{(\sigma,t)}
=
\begin{bmatrix}
0_{d\times d} & 0_{d\times 1}\\
0_{1\times d} & 0\\
a_t I_d       & 0_{d\times 1}\\
0_{d\times d} & 0_{d\times 1}\\
a_t\sigma\epsilon_m^\top & 1
\end{bmatrix}.
\]
Then the state-token query in head \(m\) is
\[
q_{N}^{(m)}
=
\bigl[
a_t(z+\sigma\epsilon_m),\,1
\bigr].
\]
For a data token \(i\le N\),
\[
\langle q_{N}^{(m)},k_i\rangle
=
\frac{t}{(1-t)^2}\langle z+\sigma\epsilon_m,x_i\rangle
-
\frac{t^2}{2(1-t)^2}\|x_i\|^2.
\]
This equals the Gaussian responsibility logit for the perturbed state \(z+\sigma\epsilon_m\), up to an additive term independent of \(i\). Hence
\[
a_{N,i}^{(m)}(X)
=
w_i(t,z+\sigma\epsilon_m).
\]
Using the same value projection \(W_V^{(t)}\), head \(m\) outputs
\[
\sum_{i=1}^N w_i(t,z+\sigma\epsilon_m)(t x_i)
=
k_t(z+\sigma\epsilon_m)
\]
in its scratch output. The multi-head output projection averages these heads:
\[
\frac{1}{M}\sum_{m=1}^M k_t(z+\sigma\epsilon_m)
=
k_{\sigma,t}(z).
\]
The same feedforward matrix \(W_{\mathrm{FF}}^{(t,h)}\), with \(k_t(z)\) replaced by \(k_{\sigma,t}(z)\), updates the state token to
\[
z+h v_{\sigma,t}(z)
=
z-\frac{h}{1-t}z+\frac{h}{t(1-t)}k_{\sigma,t}(z),
\]
and resets the scratch block.

Applying this construction at every layer \(\ell=0,\ldots,L-1\) with \(t=t_\ell\) and \(h=h_\ell\), the same induction gives a parameter choice \(\theta^\star_{\sigma\text{-}\mathrm{CFD}}\) such that
\[
\Pi_{\mathrm{state}}
\left(
\mathcal{T}_{\theta^\star_{\sigma\text{-}\mathrm{CFD}}}
\bigl(
\mathrm{Enc}(\{x_i\}_{i=1}^N,z_0)
\bigr)
\right)
=
z_L,
\]
where \(z_L\) is generated by the smoothed closed-form diffusion recursion. This completes the proof.
\section{Proof of Theorem~\ref{thm:efs-realization}}
\label{app:proof-efs-realization}

\subsection{Prompt encoding for EFS}
\label{subsec:efs-encoding}

We now describe the prompt used in Theorem~\ref{thm:efs-realization}. The EFS parameters \(\tau=(K,\gamma,s,\epsilon)\) are fixed and compiled into the transformer weights. The prompt stores only the instance-specific quantities: the initial particles \(x_1^{(0)},\ldots,x_N^{(0)}\) and the reference point \(z^{(K)}\).

We use \(m=N+2\) tokens. The first \(N\) rows are particle tokens, row \(N+1\) stores the reference point \(z^{(K)}\), and row \(N+2\) is the state token. Each token is partitioned into the following blocks:
\[
\bigl[
(x^{[0]},x^{[1]},\ldots,x^{[K]})
\;\big|\;
(r^{[0]},r^{[1]},\ldots,r^{[K]})
\;\big|\;
z
\;\big|\;
r_z
\;\big|\;
g
\;\big|\;
c
\;\big|\;
u
\;\big|\;
b
\;\big|\;
\eta
\;\big|\;
\xi
\bigr].
\]
Here \(x^{[j]}\in\mathbb{R}^d\) stores the \(j\)-th forward particle iterate and \(r^{[j]}\in\mathbb{R}\) its squared norm; these quantities are approximated during the computation. The blocks \(z\in\mathbb{R}^d\) and \(r_z\in\mathbb{R}\) store the backward point and its squared norm. The vector \(g\in\mathbb{R}^d\) accumulates repulsive interactions, \(c=1\) is constant, and \(u\in\mathbb{R}^d\), \(b\in\mathbb{R}\) are separate attention scratch registers. The bits \(\eta\) and \(\xi\) select the state and reference tokens, respectively. Therefore,
\[
p_0=(K+3)d+K+3,\qquad p=p_0+d+3=(K+4)d+K+6.
\]

The first six blocks form \(X_{\mathrm{base}}\in\mathbb{R}^{(N+2)\times p_0}\):
\[
X_{\mathrm{base}}
=
\left[
\begin{array}{c|c|c|c|c|c}
(x_1^{(0)})^\top,\;0_d^\top,\ldots,0_d^\top
&
\|x_1^{(0)}\|^2,\;0,\ldots,0
&
0_d^\top
&
0
&
0_d^\top
&
1
\\
\vdots
&
\vdots
&
\vdots
&
\vdots
&
\vdots
&
\vdots
\\
(x_N^{(0)})^\top,\;0_d^\top,\ldots,0_d^\top
&
\|x_N^{(0)}\|^2,\;0,\ldots,0
&
0_d^\top
&
0
&
0_d^\top
&
1
\\
0_d^\top,\;0_d^\top,\ldots,0_d^\top
&
0,\;0,\ldots,0
&
(z^{(K)})^\top
&
\|z^{(K)}\|^2
&
0_d^\top
&
1
\\
0_d^\top,\;0_d^\top,\ldots,0_d^\top
&
0,\;0,\ldots,0
&
0_d^\top
&
0
&
0_d^\top
&
1
\end{array}
\right]
.
\]
Writing \(e_i\) for the \(i\)-th standard basis vector in \(\mathbb{R}^{N+2}\), the complete prompt \(X_{\mathrm{EFS}}\in\mathbb{R}^{(N+2)\times p}\) is
\[
X_{\mathrm{EFS}}
=\bigl[X_{\mathrm{base}}\mid 0_{(N+2)\times d}\mid 0_{N+2}\mid e_{N+2}\mid e_{N+1}\bigr].
\]
Thus \(\eta_i=1\) only for the state token and \(\xi_i=1\) only for the reference token; \(\chi_i=1-\eta_i-\xi_i\) selects the particle tokens. These constant features are preserved exactly.

Initially, particle row \(i\le N\) stores \(x_i^{(0)}\) and \(\|x_i^{(0)}\|^2\), while the later blocks \(x^{[1]},\ldots,x^{[K]}\) and \(r^{[1]},\ldots,r^{[K]}\) are zero. During the forward phase, the transformer fills these blocks with approximations of the forward EFS iterates \(x_i^{(j)}\) and their squared norms. The reference row stores \(z^{(K)}\) and \(\|z^{(K)}\|^2\). The final row is the state token used during the backward phase. The readout extracts the \(z\)-block of this final row:
\[
\Pi_{\mathrm{state}}(X)=X_{N+2}^{(z)}.
\]
After the transformer computation, this readout approximates the EFS output \(z^{(0)}\).
All \(x^{[j]}\)- and \(r^{[j]}\)-blocks of the reference and state tokens remain exactly zero. The selectors distinguish a particle at the origin from a reference or state token. The register \(r_z\) is reserved for the squared norm and is never used as attention scratch space.

For an attention head \(h\), queries, keys, and values are
\[
q_i=X_iW_{Q,h},\qquad
k_i=X_iW_{K,h},\qquad
v_i=X_iW_{V,h}.
\]
We use standard softmax attention over the \(N+2\) tokens:
\[
\alpha_{i\ell}^{(h)}
=
\frac{\exp(\langle q_i,k_\ell\rangle)}
{\sum_{r=1}^{N+2}\exp(\langle q_i,k_r\rangle)},
\qquad
\ell=1,\ldots,N+2,
\]
and
\[
\mathrm{Attn}^{(h)}(X)_i
=
\sum_{\ell=1}^{N+2}\alpha_{i\ell}^{(h)}v_\ell .
\]

\subsection{Explicit kernel head}

Fix a stage \(j\in\{0,\ldots,K\}\) and a kernel parameter \(\lambda>0\). We first consider the forward-stage query token \(i\le N\). The goal is to recover
\[
\sum_{\ell=1}^N
e^{-\lambda\|x_i^{[j]}-x_\ell^{[j]}\|^2}
\bigl(x_i^{[j]}-x_\ell^{[j]}\bigr).
\]

Let the query/key dimension be \(d+1\). Define \(W_{Q,\lambda}^{[j]}\in\mathbb{R}^{p\times(d+1)}\) by
\[
W_{Q,\lambda}^{[j]}:
\qquad
x^{[j]}\mapsto 2\lambda x^{[j]},
\qquad
c\mapsto 1,
\]
with all other input blocks mapped to zero. Equivalently,
\[
q_i=
[2\lambda x_i^{[j]}\mid 1].
\]
Define \(W_{K,\lambda}^{[j]}\in\mathbb{R}^{p\times(d+1)}\) by
\[
W_{K,\lambda}^{[j]}:
\qquad
x^{[j]}\mapsto x^{[j]},
\qquad
r^{[j]}\mapsto -\lambda r^{[j]},
\]
with all other input blocks mapped to zero. Thus
\[
k_\ell=
[x_\ell^{[j]}\mid -\lambda r_\ell^{[j]}].
\]
First consider the ideal arithmetic computation with exact squared-norm registers; errors in these registers are treated below. For a particle token \(\ell\le N\),
\[
\langle q_i,k_\ell\rangle
=
2\lambda\langle x_i^{[j]},x_\ell^{[j]}\rangle
-
\lambda\|x_\ell^{[j]}\|^2.
\]
For both the reference token \(N+1\) and the state token \(N+2\), the \(x^{[j]}\)- and \(r^{[j]}\)-blocks are zero, so their keys and logits are zero. Consequently, \(\alpha_{i,N+1}=\alpha_{i,N+2}\).

Let \(\alpha_{i\ell}\) be the attention weight on token \(\ell\), and write \(\alpha_{i0}\equiv\alpha_{i,N+2}\) for the attention weight on the state token. Then
\[
\frac{\alpha_{i\ell}}{\alpha_{i0}}
=
\exp\!\left(
2\lambda\langle x_i^{[j]},x_\ell^{[j]}\rangle
-
\lambda r_\ell^{[j]}
\right),
\qquad \ell\le N.
\]
Multiplying by \(e^{-\lambda r_i^{[j]}}\) gives
\[
e^{-\lambda r_i^{[j]}}
\frac{\alpha_{i\ell}}{\alpha_{i0}}
=
e^{-\lambda\|x_i^{[j]}-x_\ell^{[j]}\|^2}.
\]

Now define the value projection \(W_{V,\lambda}^{[j]}\in\mathbb{R}^{p\times p}\) by
\[
W_{V,\lambda}^{[j]}:
\qquad x^{[j]}\mapsto u,\qquad \eta\mapsto b,
\]
with all other output blocks zero. Initialize \(u=b=0\) before this head. After the attention residual, these registers contain
\[
u_i=\sum_{\ell=1}^N\alpha_{i\ell}x_\ell^{[j]},
\qquad b_i=\alpha_{i,N+2}=\alpha_{i0}.
\]
The total attention mass on the particle tokens is \(1-2b_i\). Hence the continuous map
\[
\mathcal{R}_{\lambda}(y,r,u,b)
=\frac{e^{-\lambda r}}{b}\bigl((1-2b)y-u\bigr)
\]
satisfies
\[
\mathcal{R}_{\lambda}(x_i^{[j]},r_i^{[j]},u_i,b_i)
=\sum_{\ell=1}^N e^{-\lambda\|x_i^{[j]}-x_\ell^{[j]}\|^2}
(x_i^{[j]}-x_\ell^{[j]}).
\]
For fixed \(\lambda\), logits are bounded on a compact register domain, so \(b_i\) is bounded away from zero. A positionwise ReLU feedforward network can therefore approximate \(\mathcal{R}_{\lambda}\) uniformly there. It writes the interaction into the accumulator \(g\) and resets \(u,b\) to zero; the routing and accumulation are specified below.

\subsection{Backward-stage kernel head}

For the backward stage, the query token is the state token \(N+2\), whose \(z\)-block stores the current backward iterate \(z^{(j)}\). The key and value projections remain \(W_{K,\lambda}^{[j]}\) and \(W_{V,\lambda}^{[j]}\). The query projection is replaced by \(\widetilde W_{Q,\lambda}\), defined by
\[
\widetilde W_{Q,\lambda}:
\qquad
z\mapsto 2\lambda z,
\qquad
c\mapsto 1,
\]
with all other input blocks mapped to zero. Thus
\[
q_{N+2}=[2\lambda z_{N+2}\mid 1],
\qquad
k_\ell=[x_\ell^{[j]}\mid -\lambda r_\ell^{[j]}].
\]
Using \(\mathcal{R}_{\lambda}(z_{N+2},r_{z,N+2},u_{N+2},b_{N+2})\), the same argument yields, up to arbitrary uniform approximation,
\[
\sum_{\ell=1}^N
e^{-\lambda\|z_{N+2}-x_\ell^{[j]}\|^2}
\bigl(z_{N+2}-x_\ell^{[j]}\bigr).
\]

\subsection{Approximating the EFS interaction field}

Let \(a=s/2+1>0\). Since \(\epsilon>0\), the Laplace identity gives
\[
\frac{v}{(\|v\|^2+\epsilon)^a}
=
\frac{1}{\Gamma(a)}
\int_0^\infty
t^{a-1}e^{-t\epsilon}e^{-t\|v\|^2}v\,dt .
\]
On the compact domain, all relevant differences satisfy \(\|v\|\le R\). Hence, for every \(\delta>0\), there exist \(H\in\mathbb{N}\), nodes \(\lambda_1,\ldots,\lambda_H>0\), and coefficients \(\omega_1,\ldots,\omega_H>0\) such that
\[
\sup_{\|v\|\le R}
\left\|
\frac{v}{(\|v\|^2+\epsilon)^a}
-
\sum_{h=1}^H
\omega_h e^{-\lambda_h\|v\|^2}v
\right\|
\le \delta .
\]
Indeed, the integrand norm is bounded by \(R t^{a-1}e^{-\epsilon t}/\Gamma(a)\), an integrable function. Truncating the integral near zero and infinity and taking a finite Riemann sum on the remaining interval gives the stated uniform approximation.

We evaluate the \(H\) kernel heads \emph{sequentially in separate blocks}. Starting from \(g=u=b=0\), the feedforward map after head \(h\) adds \(\omega_h\mathcal{R}_{\lambda_h}(y,r,u,b)\), up to uniform approximation, to \(g\) on the active tokens and resets \(u,b\) exactly. The active selector is \(\chi\) during the forward phase and \(\eta\) during the backward phase. Thus each head's nonlinear normalization is evaluated before summing its contribution. At the end, \(g\) approximates
\[
\sum_{\ell=1}^N
\frac{y-x_\ell^{[j]}}
{(\|y-x_\ell^{[j]}\|^2+\epsilon)^{s/2+1}}
\]
up to uniform error.

The attractive term is affine after summation:
\[
\sum_{\ell=1}^N(y-x_\ell^{[j]})
=
Ny-\sum_{\ell=1}^N x_\ell^{[j]}.
\]
It is computed by a uniform-attention head. Explicitly, choose
\[
W_Q^{\mathrm{avg}}=0,\qquad
W_K^{\mathrm{avg}}=0,
\qquad
W_V^{\mathrm{avg}}:\; x^{[j]}\mapsto u,
\]
with all other output blocks zero. Then all logits are zero, so attention returns
\[
\mathrm{Attn}^{\mathrm{avg}}(X)_i^{(u)}
=
\frac{1}{N+2}\sum_{\ell=1}^{N+2}x_\ell^{[j]}
=
\frac{1}{N+2}\sum_{\ell=1}^{N}x_\ell^{[j]},
\]
because the reference and state tokens have zero \(x^{[j]}\)-blocks. This head writes into the cleared register \(u\), leaving \(g\) unchanged. Combining the two registers gives the force approximation
\[
\widehat F(y)=c_N\bigl(Ny-(N+2)u-g\bigr),
\qquad c_N=\frac{2}{N(N-1)},
\]
to the normalized field
\[
F_{N,\epsilon}(y;x_1,\ldots,x_N)
=c_N\sum_{\ell=1}^N\left(y-x_\ell-
\frac{y-x_\ell}{(\|y-x_\ell\|^2+\epsilon)^{s/2+1}}\right).
\]
This is \(\nabla_{x_i}E_{N,\epsilon}\) at \(y=x_i\) and \(\nabla_z V_{N,\epsilon}\) at \(y=z\). The term with \(x_\ell=y\) is zero.

\subsection{Forward and backward simulation}

\paragraph{Exact routing and scratch reset.}
The feedforward network is shared across tokens. It can nevertheless restrict an approximate continuous update to tokens selected by a fixed bit \(t\in\{0,1\}\). To see this, write a two-layer ReLU approximant in column notation as \(A\operatorname{ReLU}(Bw+a)+d\). On the compact register domain, choose \(M\) larger than every component of \(Bw+a\). Then
\[
A\operatorname{ReLU}\bigl(Bw+a-M(1-t)\mathbf{1}\bigr)+dt
\]
equals the approximant when \(t=1\) and is exactly zero when \(t=0\). The affine term \(dt\) is also realizable by ReLU units. We use \(t=\chi\) or \(t=\eta\), as appropriate. Immutable output coordinates are identically zero in the feedforward residual; scratch reset uses the exact linear residual \((u,b)\mapsto(-u,-b)\), and similarly for \(g\) at the end of a stage. Thus inactive trajectory registers and selectors remain unchanged exactly, including the zero particle-history blocks of the reference and state tokens.

\paragraph{Forward stages.}
At stage \(j=0,\ldots,K-1\), the \(H\) sequential kernel blocks followed by the uniform-attention block compute \(\widehat F(x_i^{[j]})\). The final feedforward map, gated by \(\chi_i\), writes into the previously zero future blocks
\[
x_i^{[j+1]}=x_i^{[j]}-\gamma\widehat F(x_i^{[j]}),
\qquad r_i^{[j+1]}\approx\|x_i^{[j+1]}\|^2,
\qquad i\le N,
\]
and resets \(g,u,b\) on every token. The coordinate update is affine in the available registers; the squared-norm update is approximated jointly with it by the feedforward network. Earlier trajectory blocks are preserved. The particle and state \(z,r_z\)-blocks remain zero throughout the forward phase, while the reference \(z,r_z\)-blocks remain unchanged.

\paragraph{Copying the reference point.}
After the forward phase, use \(W_Q=W_K=0\) and a value map \(z\mapsto u\), \(r_z\mapsto b\). Only the reference row can have nonzero \(z,r_z\), so the attention residual gives
\[
u_i=\frac{z^{(K)}}{N+2},\qquad b_i=\frac{\|z^{(K)}\|^2}{N+2}.
\]
An \(\eta_i\)-gated affine feedforward map writes \((N+2)u_i,(N+2)b_i\) into the state token's initially zero \(z,r_z\)-blocks and resets \(u,b\). This copy is exact on the bounded domain and leaves the reference token unchanged.

\paragraph{Backward stages.}
For \(j=K,K-1,\ldots,1\), use the backward query, the stored \(j\)-th particle trajectory, and the selector \(\eta\). The same sequential construction approximates \(F_{N,\epsilon}(z;x_1^{[j]},\ldots,x_N^{[j]})\). The final feedforward map updates only the state token:
\[
z\leftarrow z+\gamma\widehat F(z),
\qquad r_z\approx\|z+\gamma\widehat F(z)\|^2,
\]
where the right-hand sides use the old registers. All trajectory and reference registers are preserved, and \(g,u,b\) are reset on every token.

\paragraph{Uniform error control.}
First compare the exact recurrence with the ideal quadrature computation, using exact squared norms. On a compact neighborhood of all admissible trajectories with positive margin, the \(2K\) exact stage maps, including particle histories, have a common Lipschitz bound \(L_0\ge1\) in the maximum of the vector-register Euclidean norms, since \(\epsilon>0\). A quadrature error \(\delta\) per pair gives force error at most \(c_NN\delta\), so
\[
d_{q+1}\le L_0d_q+\gamma c_NN\delta,\qquad d_0=0.
\]
Choose \(\delta\) so that
\[
\gamma c_NN\delta\sum_{q=0}^{2K-1}L_0^q\le\varepsilon/2
\]
and the left-hand side is smaller than half the neighborhood margin. Choose the quadrature after fixing \(\delta\). Induction keeps its ideal trajectories in this neighborhood and gives terminal error at most \(\varepsilon/2\).

For the fixed quadrature, the ideal program has \(J\) locally Lipschitz blocks and compact intermediate register sets; kernel attention weights \(b\) are bounded away from zero. Choose a common neighborhood margin \(\rho>0\) and block Lipschitz bound \(C\ge1\). Use the maximum, over all tokens and registers, of Euclidean vector norms and absolute scalar values. Approximating each nonlinear feedforward map within \(\kappa\), with exact routing and resets, gives
\[
e_{q+1}\le Ce_q+\kappa,\qquad e_0=0.
\]
Choose \(\kappa\) such that
\[
\kappa\sum_{q=0}^{J-1}C^q\le\min\{\rho/2,\varepsilon/2\}.
\]
This induction accounts for errors in every stored squared norm, attention logit, interaction accumulator, and state update. Inactive history blocks and scratch resets remain exact. The triangle inequality now gives
\[
\left\|
\Pi_{\mathrm{state}}\left(
\mathcal{T}_{\theta^\star_{\mathrm{EFS}}}
\bigl(\mathrm{Enc}_{\mathrm{EFS}}(x_1^{(0)},\ldots,x_N^{(0)},z^{(K)})\bigr)
\right)-z^{(0)}
\right\|\le\varepsilon.
\]
All choices depend only on the fixed parameters, compact domain, and tolerance. The construction uses \(J=2K(H+1)+1\) blocks and token width \(p=(K+4)d+K+6\), with sufficiently large finite feedforward hidden widths. This proves Theorem~\ref{thm:efs-realization}.

\newpage
\section{Experiments Details}\label{app:experimental-details}

\subsection{Smile Experiment}
\label{sec:Smile_Experiment}

Figure~\ref{fig:in-context-sampling} provided a controlled visual example of the sampling behavior studied in this work. We construct a two-dimensional smiling emoji distribution with four components: a noisy circular face boundary, two Gaussian eye clusters, and a crescent-shaped smile. The smile component is removed from the training distribution. We use a small GPT-2-style decoder-only transformer trained from scratch with 16 layers and 32-dimensional hidden embeddings. Each training example contains 64 input points, and the model is trained on 1,000 IID training sequences for 2,000 optimization steps.

Because the architecture is decoder-only and causal, we formulate the task in the same way as next-token prediction. For each IID sequence $(x_1,\ldots,x_{T+1})$, the model receives $(x_1,\ldots,x_T)$ as input and is trained to predict the shifted sequence $(x_2,\ldots,x_{T+1})$. The final target point $x_{T+1}$ is not included in the input context, so it acts as a fresh IID sample from the same distribution. Since the samples are unordered, we train with a full-sequence optimal-transport matching loss rather than an index-wise regression loss. This encourages the predicted point cloud to match the target distribution, while remaining compatible with the next-token structure of the transformer.

For visualization, we plot only the prediction at the last position. It is produced after the model has attended to the entire in-context sequence. Earlier positions are also trained, but each of them is conditioned only on a shorter prefix because of the causal mask. The held-out smile gives the central qualitative test. Although smile-shaped samples are never included in the training distribution, the model can generate new IID samples along the same smile geometry.

\subsection{Synthetic layerwise geometry}
\label{app:layerwise_geometry}

\begin{figure}[h!]
    \centering
    \includegraphics[width=0.95\linewidth]{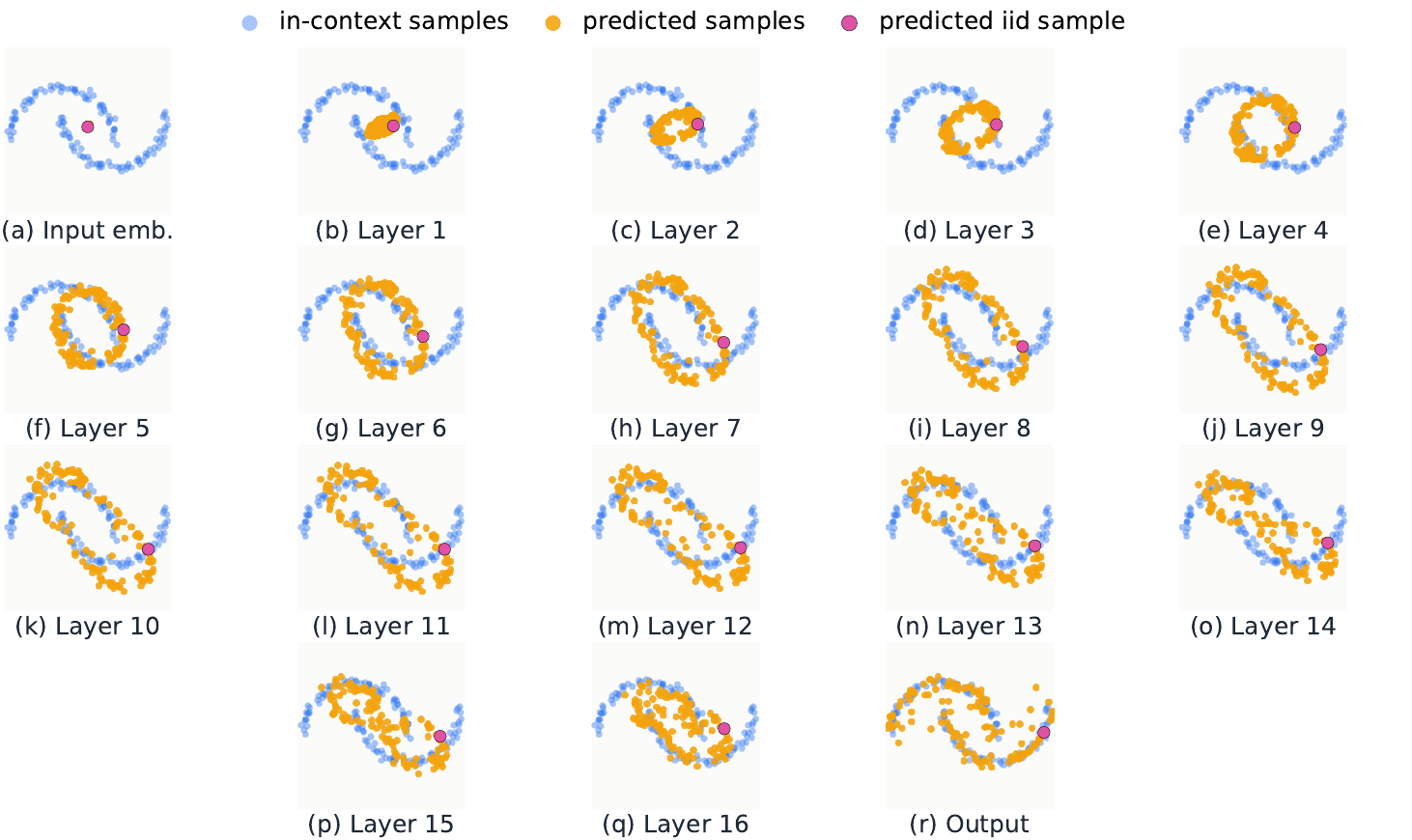}
    \caption{
    Geometry evolution of one test sample across the transformer layers. 
    }
    \label{fig:full-layer-geometry}
\end{figure}

Figure~\ref{fig:geometry-evolution} used the two-moons distribution to show how the learned sampler transforms an in-context point cloud across depth. We generate a pool of 5,000 2-dimensional points and split it at the point level. Using the same causal next-token sampling setup as in Subsection~\ref{sec:Smile_Experiment}, we train the model for 2,000 optimization steps on 1,000 IID training sequences, where each input sequence contains 64 points. To inspect the internal computation, we project the hidden states after selected transformer layers back to the two-dimensional output space using the model's output head. In early layers, the predicted points are concentrated in a small region, then expand in the middle layers toward a more uniform geometry, and finally contract back onto the structured two-moons distribution specified by the in-context samples. This layerwise trajectory visually supports the view of transformer depth as an iterative transport process. Full layer-by-layer visualization is provided in Fig~\ref{fig:full-layer-geometry}.

\subsection{Layerwise Uniformization in Large Language Models}
\label{app:uniformization_models_results}
\begin{figure*}[h]
    \centering

    \begin{subfigure}[t]{0.32\textwidth}
        \centering
        \includegraphics[width=\linewidth]{imgs/iid_topics/meta-llama_Llama-3.3-70B-Instruct_MMD2_distance_to_uniform_iid.pdf}
        \caption{Llama-3.3-70B-Instruct~\cite{grattafiori2024llama}}
        \label{fig:iid-large-qwen32b}
    \end{subfigure}
    \hfill
    \begin{subfigure}[t]{0.32\textwidth}
        \centering
        \includegraphics[width=\linewidth]{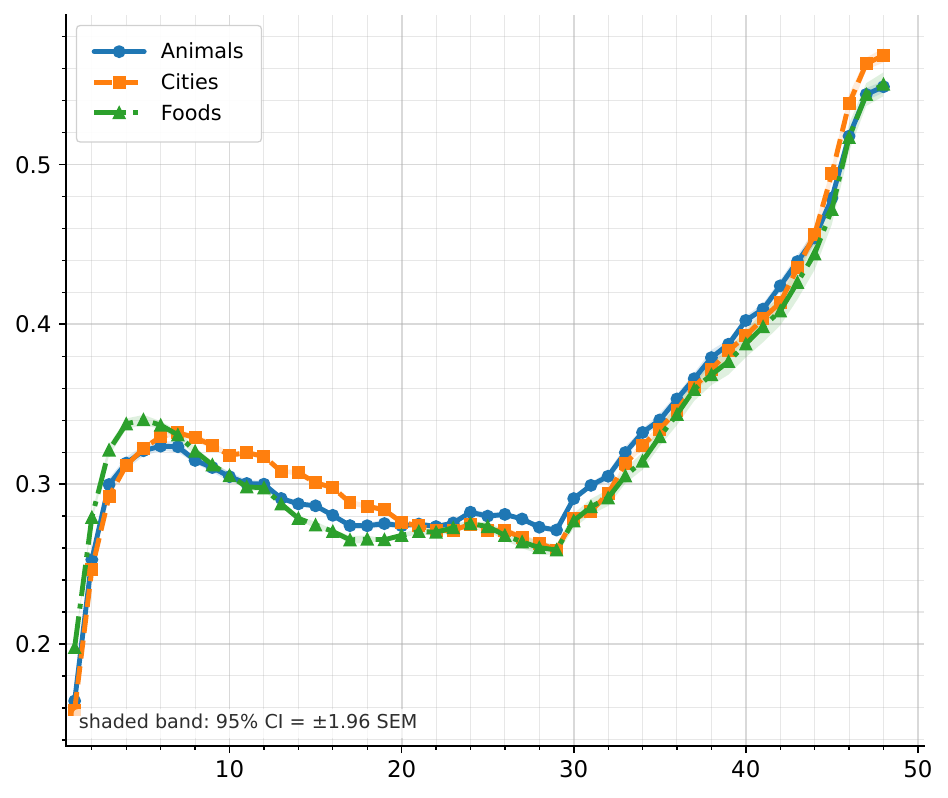}
        \caption{GPT-2 XL~\cite{radford2019language}}
        \label{fig:iid-large-gpt2xl}
    \end{subfigure}
    \hfill
    \begin{subfigure}[t]{0.32\textwidth}
        \centering
        \includegraphics[width=\linewidth]{imgs/iid_topics/Qwen_Qwen2.5-14B_MMD2_distance_to_uniform_iid.pdf}
        \caption{Qwen2.5-14B~\cite{bai2023qwen}}
        \label{fig:iid-large-opt66b}
    \end{subfigure}
    
    \begin{subfigure}[t]{0.32\textwidth}
        \centering
        \includegraphics[width=\linewidth]{imgs/iid_topics/Qwen_Qwen2.5-32B_MMD2_distance_to_uniform_iid.pdf}
        \caption{Qwen2.5-32B~\cite{bai2023qwen}}
        \label{fig:iid-large-llama70b}
    \end{subfigure}
\hspace{0.002\textwidth}
    \begin{subfigure}[t]{0.32\textwidth}
        \centering
        \includegraphics[width=\linewidth]{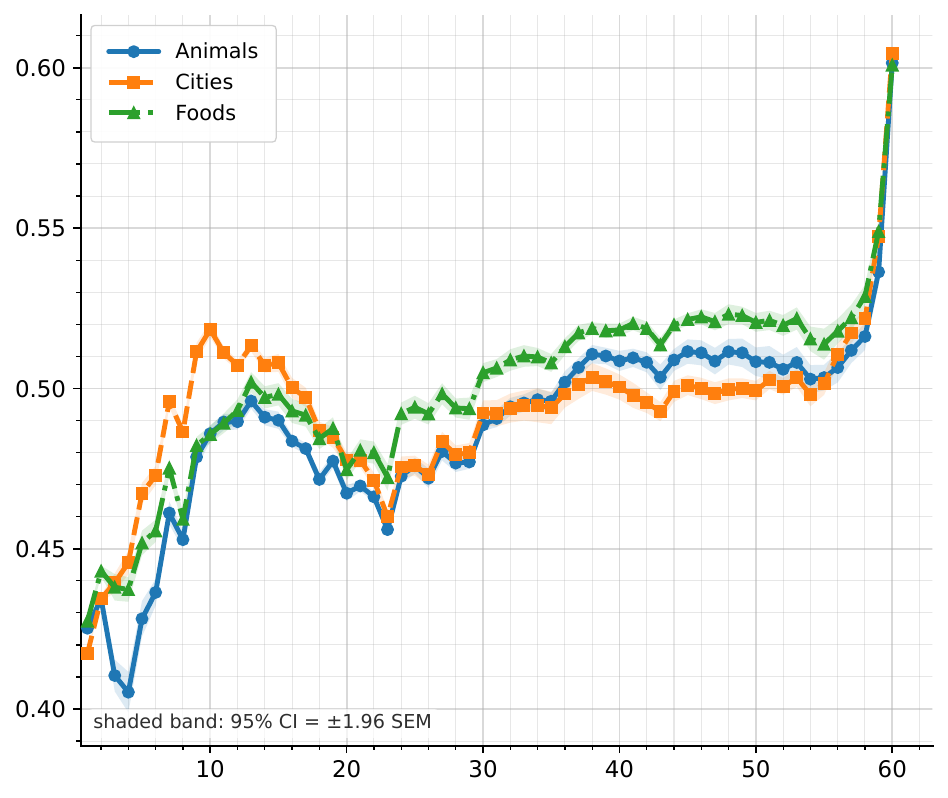}
        \caption{Falcon-40B~\cite{almazrouei2023falcon}}
        \label{fig:iid-large-falcon40b}
    \end{subfigure}
\hspace{0.002\textwidth}
    \begin{subfigure}[t]{0.32\textwidth}
        \centering
        \includegraphics[width=\linewidth]{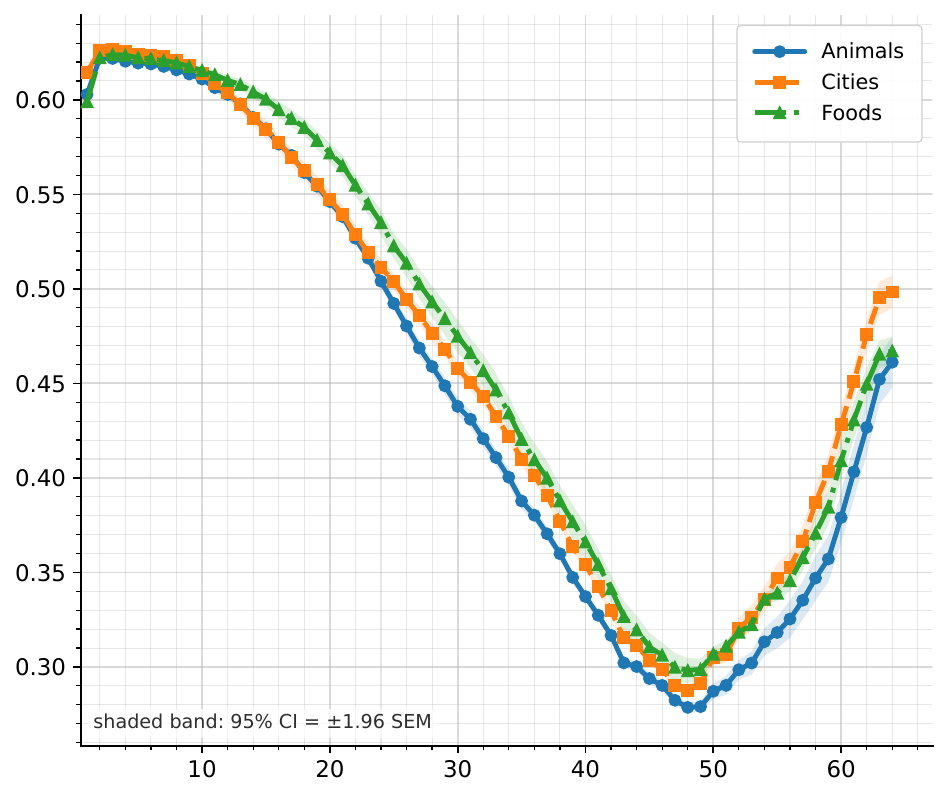}
        \caption{OPT-66B~\cite{zhang2022opt}}
        \label{fig:iid-large-opt66b-repeat}
    \end{subfigure}

    \begin{subfigure}[t]{0.32\textwidth}
        \centering
        \includegraphics[width=\linewidth]{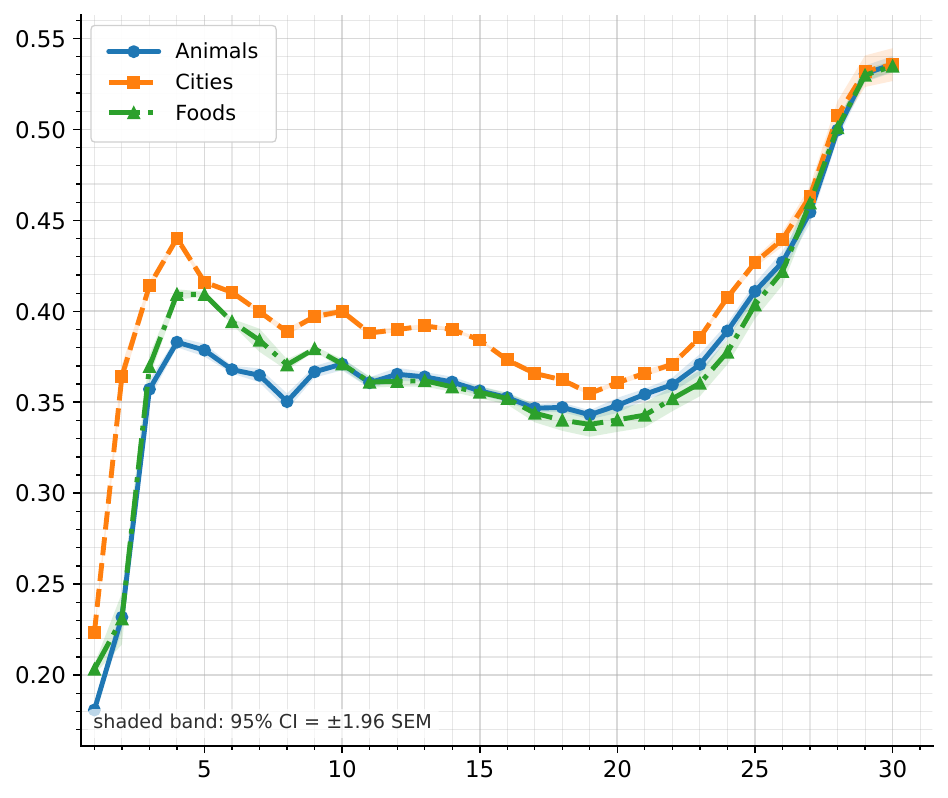}
        \caption{Bloom-7b1~\cite{workshop2022bloom}}
        \label{fig:iid-large-llama70b}
    \end{subfigure}
\hspace{0.002\textwidth}
    \begin{subfigure}[t]{0.32\textwidth}
        \centering
        \includegraphics[width=\linewidth]{imgs/iid_topics/cerebras_Cerebras-GPT-13B_MMD2_distance_to_uniform_iid.pdf}
        \caption{Cerebras-GPT-13B~\cite{dey2023cerebras}}
        \label{fig:iid-large-falcon40b}
    \end{subfigure}
\hspace{0.002\textwidth}
    \begin{subfigure}[t]{0.32\textwidth}
        \centering
        \includegraphics[width=\linewidth]{imgs/iid_topics/openlm-research_open_llama_13b_MMD2_distance_to_uniform_iid.pdf}
        \caption{Openlm-Llama-13b~\cite{openlm2023openllama}}
        \label{fig:iid-large-opt66b-repeat}
    \end{subfigure}

    \caption{
    Additional large-model results for semantic-category prompts sampled without replacement. The curves show the layerwise \(\mathrm{MMD}^2\) distance to the uniform spherical reference distribution.
    }
    \label{fig:nine_plots1b}
\end{figure*}

Figure~\ref{fig:mmd-uniform-iid} shows that pretrained autoregressive language models reshape in-context semantic distributions across depth. We construct prompts from three topic vocabularies: animals, foods, and cities. For each model, we keep only tokenizer-eligible single-token words and use unique tokens, so each prompt contains up to 256 distinct words from one topic. We run 5 independent prompts per topic and record the raw hidden states after each transformer block, excluding the embedding layer, final layer normalization, and vocabulary logits. The complete semantic-token experiment was conducted across nine pretrained autoregressive language models; the corresponding layerwise MMD results are reported in Figure~\ref{fig:nine_plots1b}.

In Figure~\ref{fig:mmd-cbt-stories}, we also repeat the same analysis on natural text prompts from the CBT dataset~\cite{hill2015goldilocks}: for each story, we feed the first 256 tokens to the model, but compute the layerwise metrics only on the first occurrence of each unique token.

At each layer, we treat the hidden states as an empirical particle cloud. We apply row-center normalization by subtracting the coordinate mean of each hidden vector and projecting it to the unit sphere. We then compare the resulting cloud to samples from a uniform spherical reference using the off-diagonal RBF \(\mathrm{MMD}^2\) statistic. We use this statistic descriptively, without an i.i.d.-based unbiasedness claim for the dependent hidden states. Row centering followed by normalization places each nonzero centered vector in \(\mathbf{1}^{\perp}\cap S^{d-1}\), a sphere of dimension \(d-2\); a uniform reference on the ambient sphere has different support. An exact uniformity interpretation would require a reference matched to this subspace. Accordingly, the reported curves are relative finite-cloud diagnostics, not evidence of equality with an ambient uniform distribution. The curves report the mean over trials, with shaded bands denoting 95\% confidence intervals.

Across several large language models, the curves show a U-shaped profile: the measured discrepancy to the chosen spherical reference decreases in middle layers and increases in later layers. In Figure~\ref{fig:energy-function-iid}, we further measure the corresponding EFS-style logarithmic interaction energy. For this metric, the normalized hidden states are rescaled to radius \(0.78\), and the energy is computed directly on the layerwise particle cloud. The energy curves support the same interpretation as the MMD plots: transformer layers first regularize the in-context distribution toward a uniform-like configuration and later recover structured semantic geometry.

\subsection{Failure mode}
\label{app:failure_mode}

\begin{figure*}[h]
    \centering

    \begin{subfigure}[t]{0.24\textwidth}
        \centering
        \includegraphics[width=\linewidth]{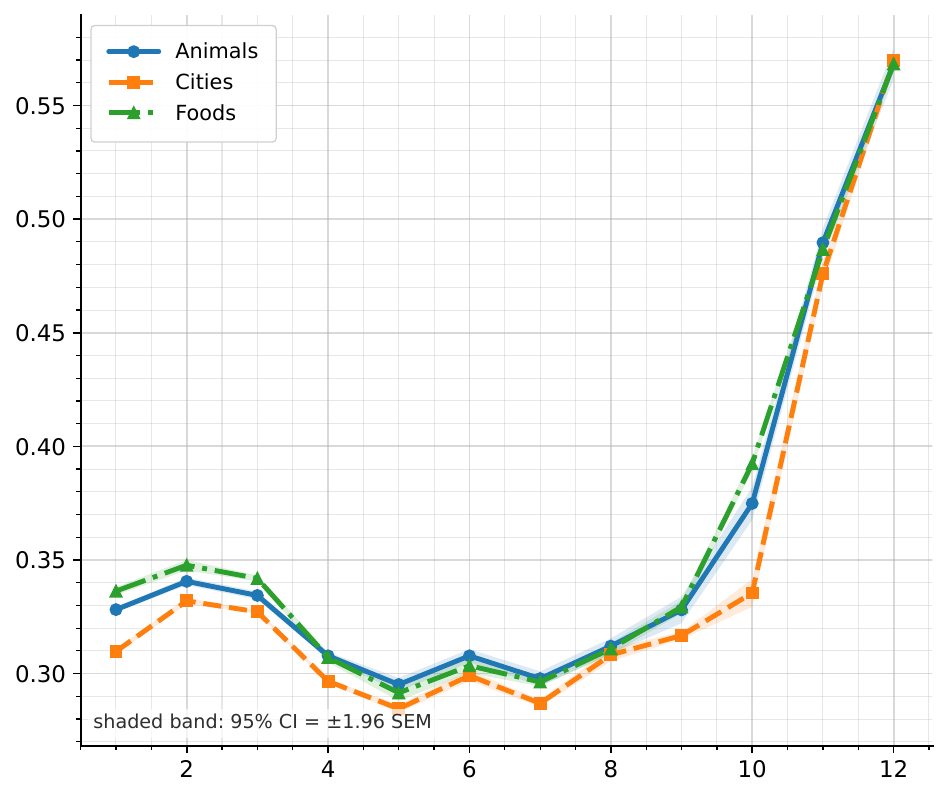}
        \caption{GPT-2}
        \label{fig:iid-gpt2-family-small}
    \end{subfigure}
    \hfill
    \begin{subfigure}[t]{0.24\textwidth}
        \centering
        \includegraphics[width=\linewidth]{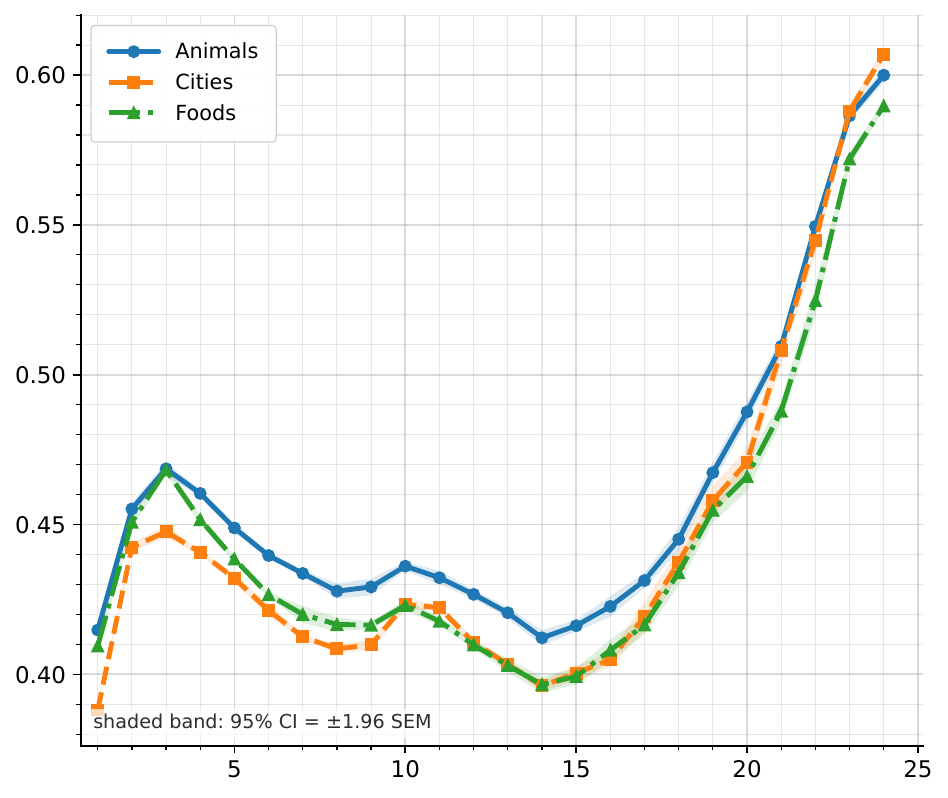}
        \caption{GPT-2 Medium}
        \label{fig:iid-gpt2-family-medium}
    \end{subfigure}
    \hfill
    \begin{subfigure}[t]{0.24\textwidth}
        \centering
        \includegraphics[width=\linewidth]{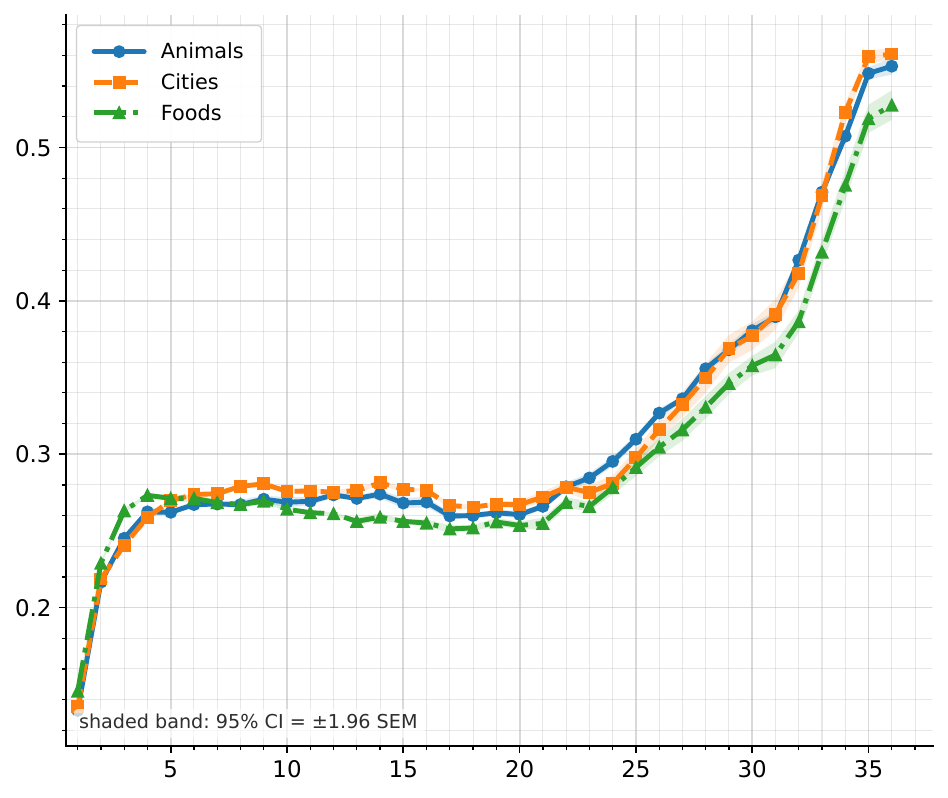}
        \caption{GPT-2 Large}
        \label{fig:iid-gpt2-family-large}
    \end{subfigure}
    \hfill
    \begin{subfigure}[t]{0.24\textwidth}
        \centering
        \includegraphics[width=\linewidth]{imgs/iid_topics/gpt2-xl_MMD2_distance_to_uniform_iid.pdf}
        \caption{GPT-2 XL}
        \label{fig:iid-gpt2-family-xl}
    \end{subfigure}
    
    \caption{
    GPT-2 family~\cite{radford2019language} on semantic-category prompts sampled without replacement. Across model sizes, the layerwise distance to the uniform spherical reference distribution decreases in the middle layers and increases near the output layers.
    }
    \label{fig:gpt2-iid-fail}
\end{figure*}

\begin{figure*}[h]
    \centering

    \begin{subfigure}[t]{0.24\textwidth}
        \centering
        \includegraphics[width=\linewidth]{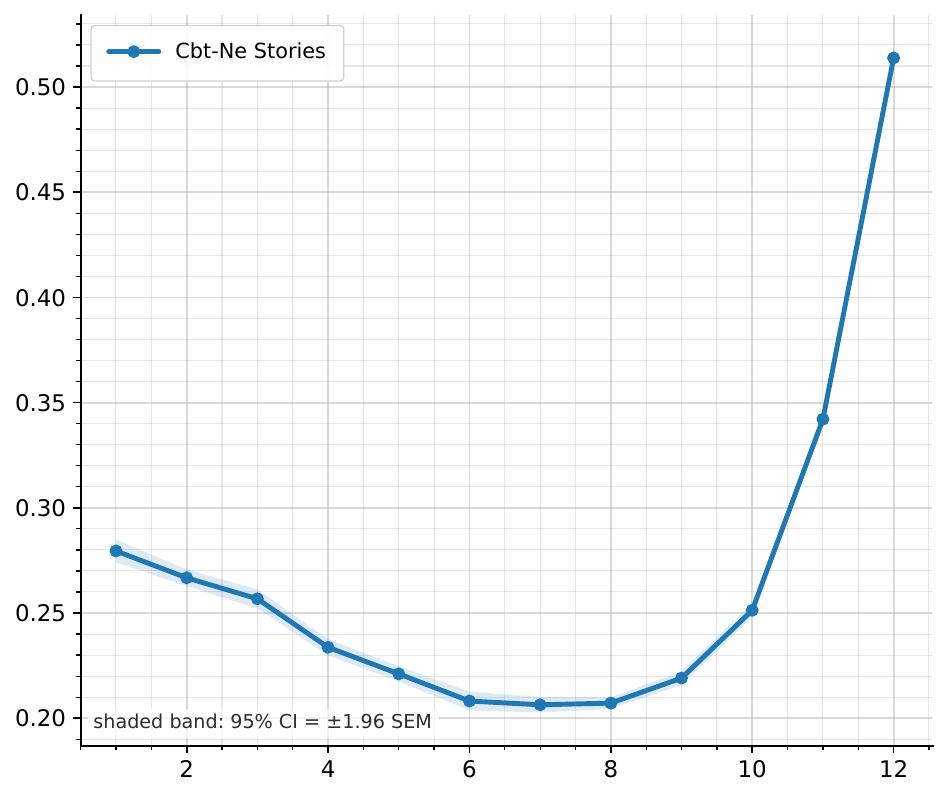}
        \caption{GPT-2}
        \label{fig:cbt-gpt2-family-small}
    \end{subfigure}
    \hfill
    \begin{subfigure}[t]{0.24\textwidth}
        \centering
        \includegraphics[width=\linewidth]{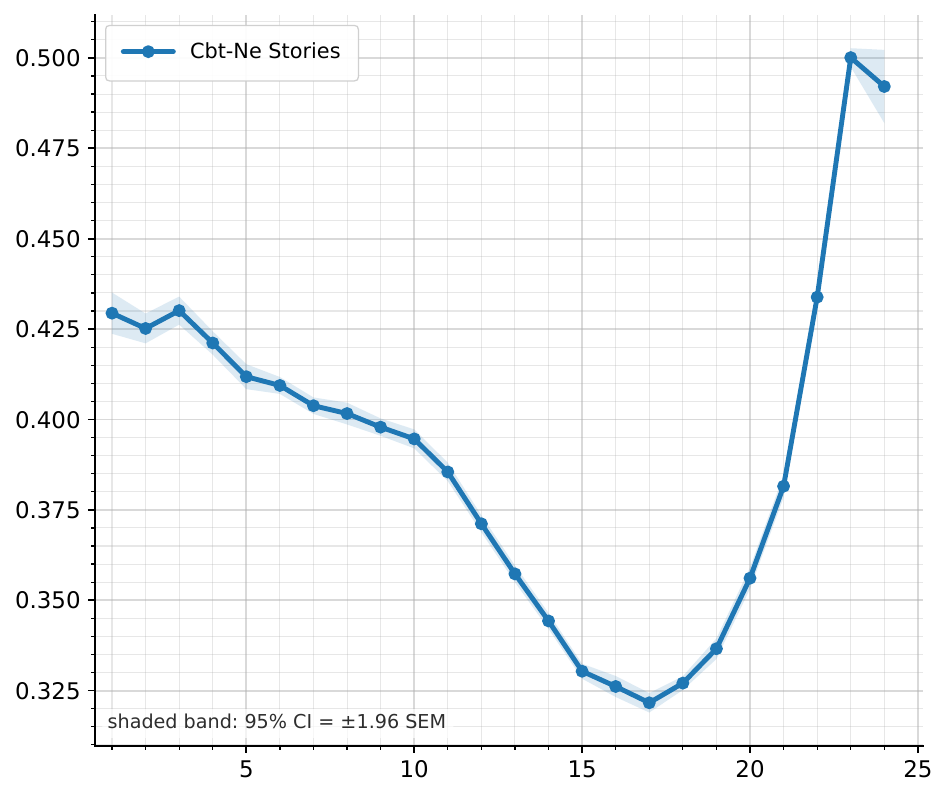}
        \caption{GPT-2 Medium}
        \label{fig:cbt-gpt2-family-medium}
    \end{subfigure}
    \hfill
    \begin{subfigure}[t]{0.24\textwidth}
        \centering
        \includegraphics[width=\linewidth]{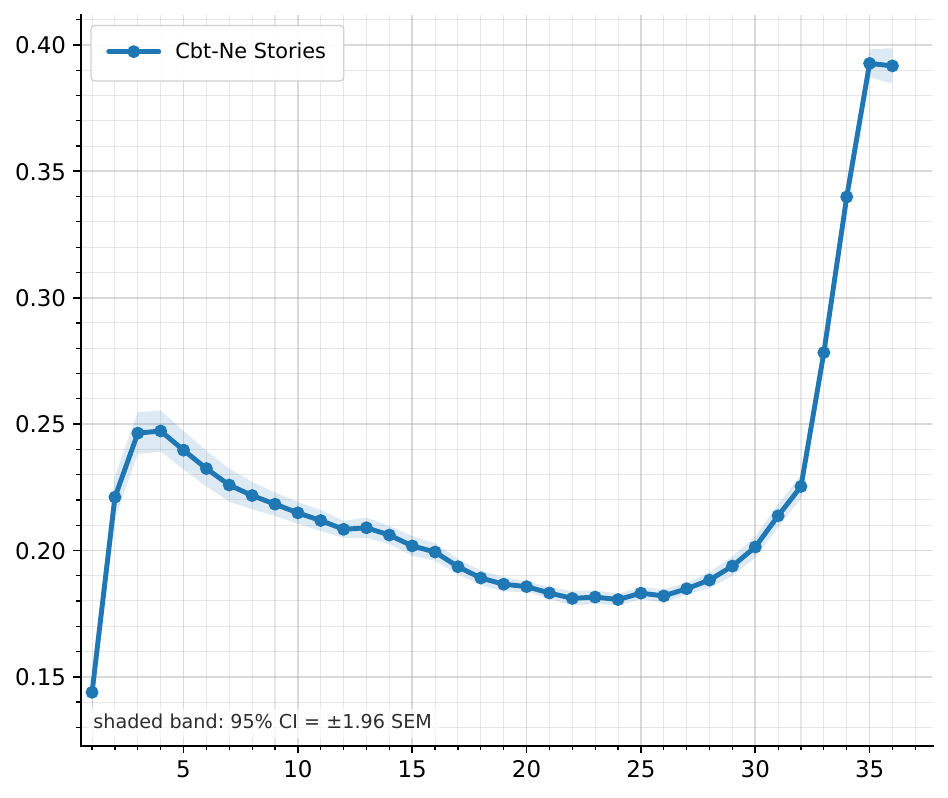}
        \caption{GPT-2 Large}
        \label{fig:cbt-gpt2-family-large}
    \end{subfigure}
    \hfill
    \begin{subfigure}[t]{0.24\textwidth}
        \centering
        \includegraphics[width=\linewidth]{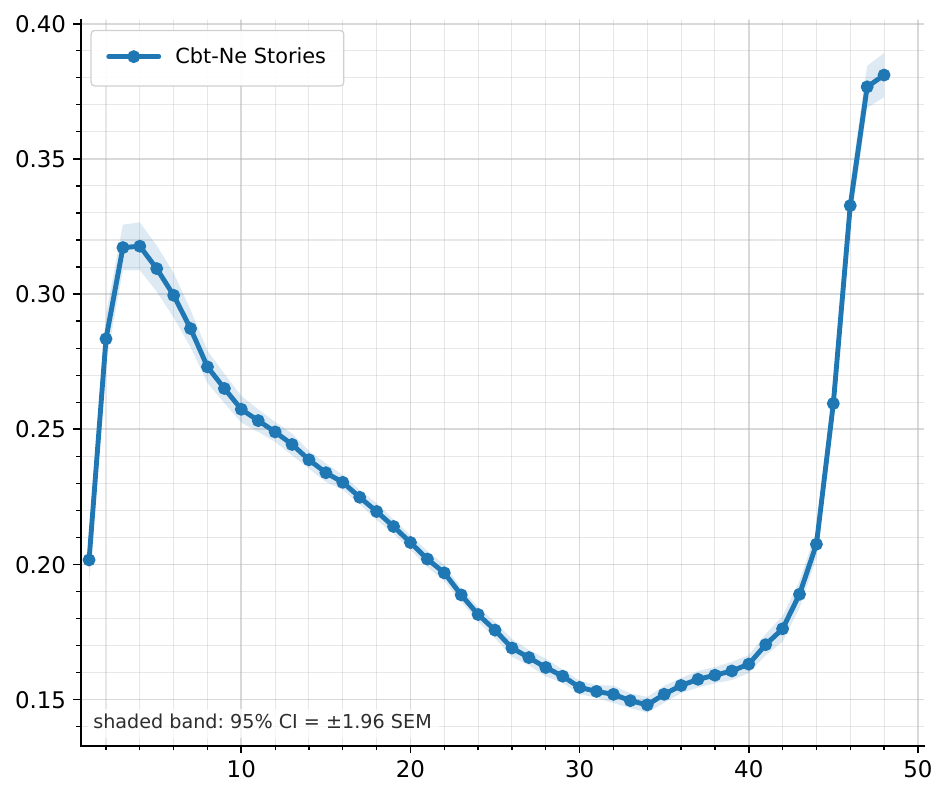}
        \caption{GPT-2 XL}
        \label{fig:cbt-gpt2-family-xl}
    \end{subfigure}
    
    \caption{
    GPT-2 family~\cite{radford2019language} on CBT story-token prompts. The same middle-layer movement toward a uniform spherical reference distribution appears for natural text prompts.
    }
    \label{fig:gpt2-cbt-fail}
\end{figure*}

\begin{figure*}[h]
    \centering

    \begin{subfigure}[t]{0.19\textwidth}
        \centering
        \includegraphics[width=\linewidth]{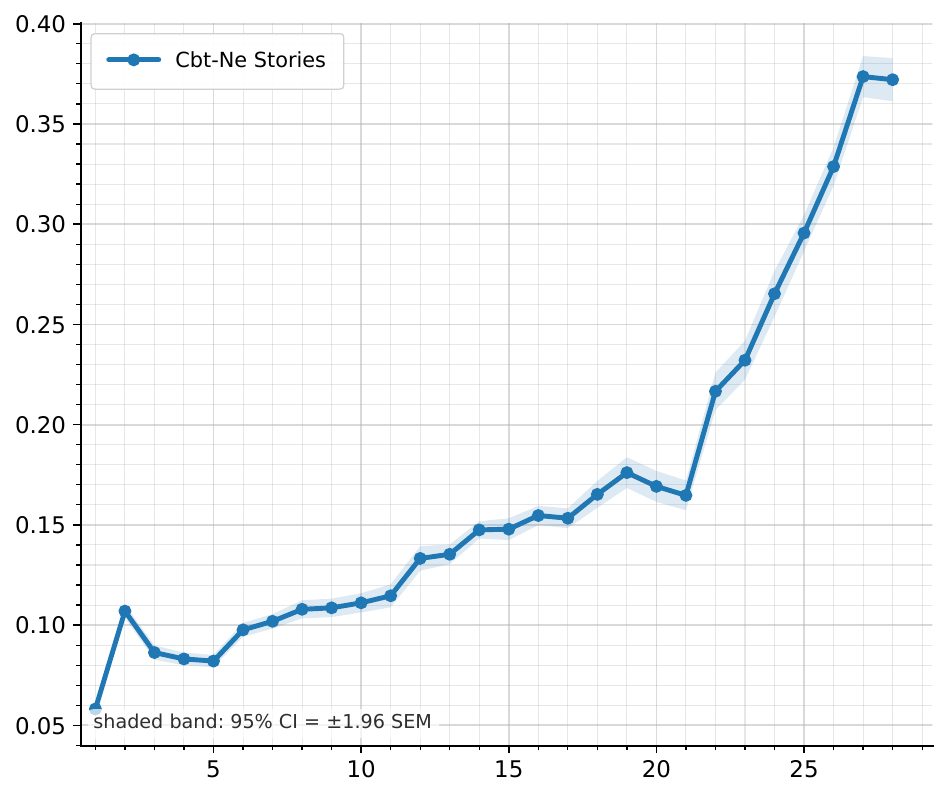}
        \caption{1.5B}
        \label{fig:qwen-family-1p5b}
    \end{subfigure}
    \hfill
    \begin{subfigure}[t]{0.19\textwidth}
        \centering
        \includegraphics[width=\linewidth]{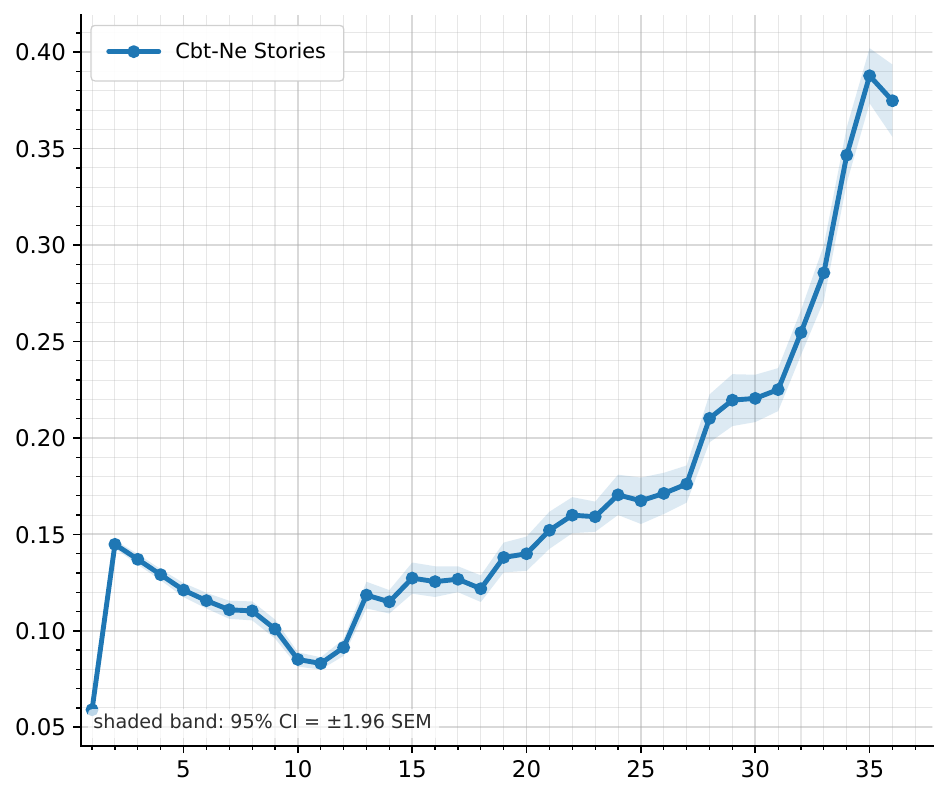}
        \caption{3B}
        \label{fig:qwen-family-3b}
    \end{subfigure}
    \hfill
    \begin{subfigure}[t]{0.19\textwidth}
        \centering
        \includegraphics[width=\linewidth]{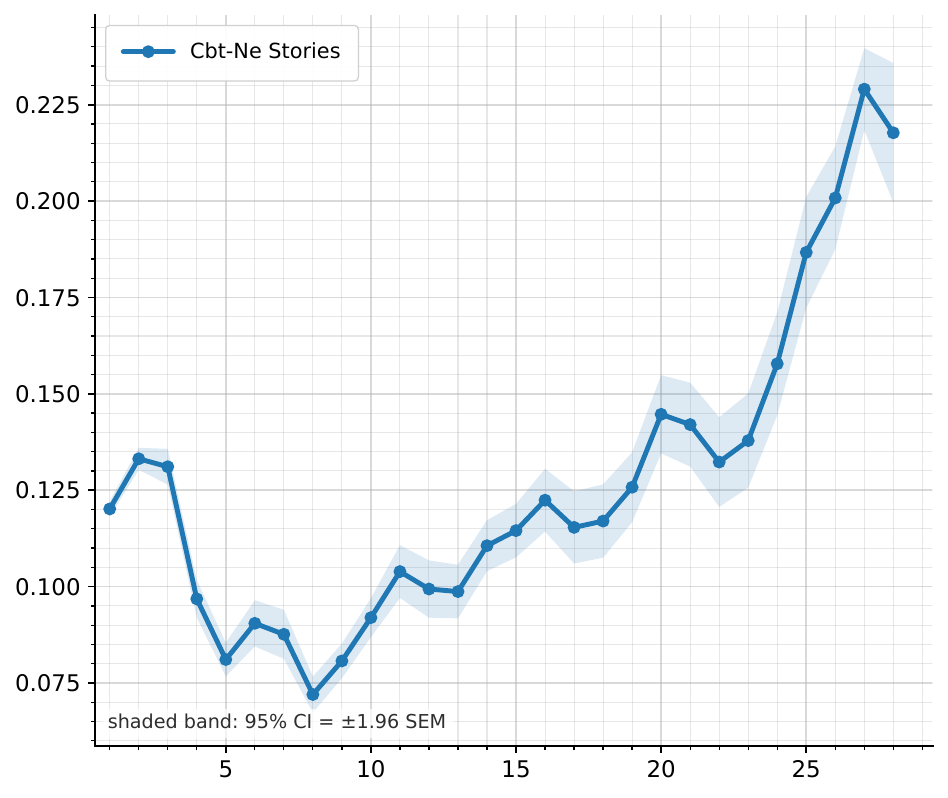}
        \caption{7B}
        \label{fig:qwen-family-7b}
    \end{subfigure}
    \hfill
    \begin{subfigure}[t]{0.19\textwidth}
        \centering
        \includegraphics[width=\linewidth]{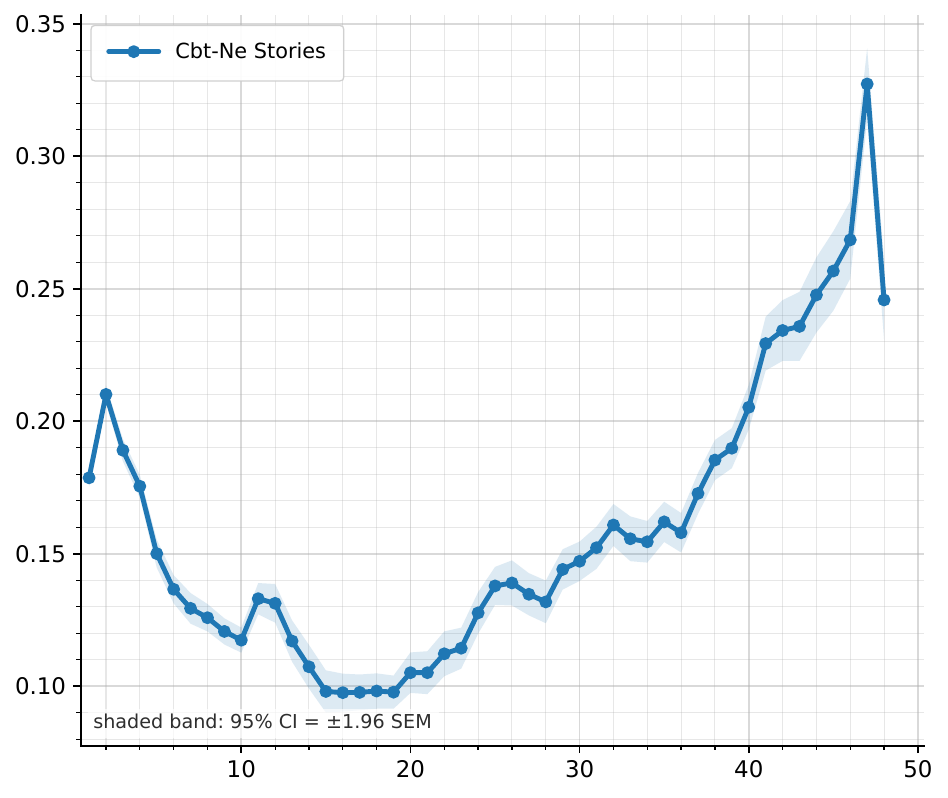}
        \caption{14B}
        \label{fig:qwen-family-14b}
    \end{subfigure}
    \hfill
    \begin{subfigure}[t]{0.19\textwidth}
        \centering
        \includegraphics[width=\linewidth]{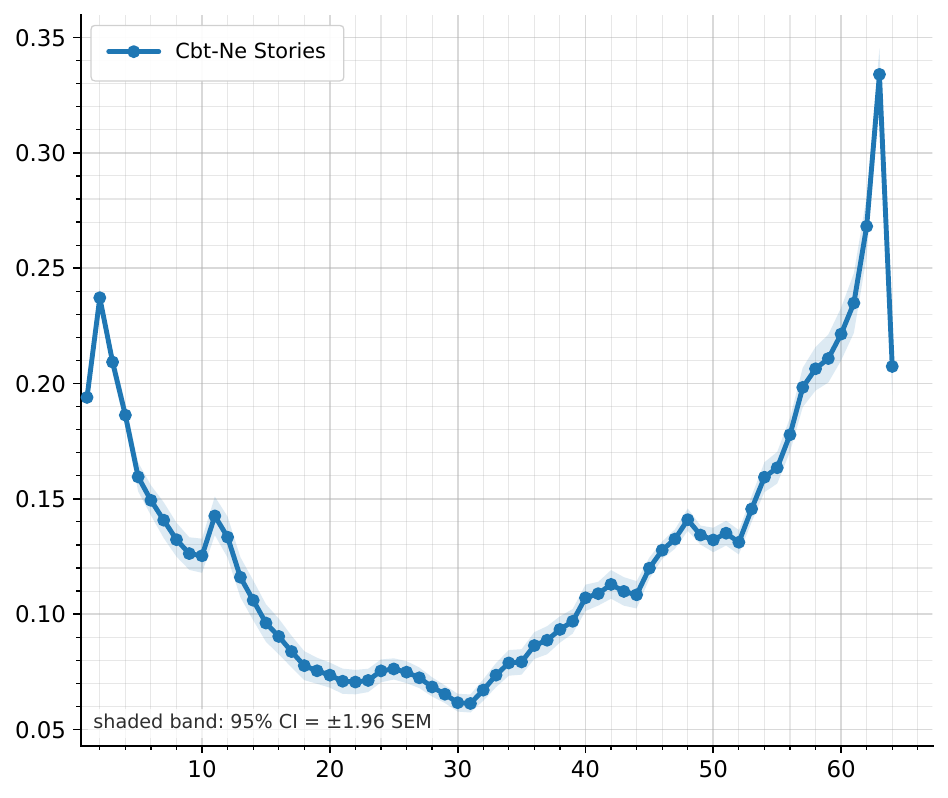}
        \caption{32B}
        \label{fig:qwen-family-32b}
    \end{subfigure}
    
    \caption{
    Qwen2.5 family~\cite{qwen2024qwen2} on CBT story-token prompts. The U-shaped profile becomes visible across model scales, indicating a middle-layer movement toward a uniform spherical reference distribution followed by a later movement away from it.
    }
    \label{fig:qwen-cbt-fail}
\end{figure*}

The layerwise uniformization pattern is not equally strong across all models. 
This effect is visible not only within the Qwen2.5 family~\cite{qwen2024qwen2}, but also within the GPT-2 family~\cite{radford2019language}, and it appears in both the semantic-topic setting and the CBT story setting; see  Figure~\ref{fig:gpt2-iid-fail}, Figure~\ref{fig:gpt2-cbt-fail}, Figure~\ref{fig:fail}, and Figure~\ref{fig:qwen-cbt-fail}. Figure~\ref{fig:fail} shows this effect within the Qwen2.5 model family ~\cite{qwen2024qwen2}. Smaller models, especially the 1.5B, 3B, and 7B variants, show only a short or weak movement toward the uniform spherical reference before the \(\mathrm{MMD}^2\) distance increases again. In these models, the middle-layer uniformization phase is compressed, suggesting that the model may not have enough depth to fully realize the two-stage sampler-like computation. The larger 14B and 32B models show a clearer separation between the early movement toward uniformity and the later movement away from it.

This behavior is consistent with the theoretical construction. In Theorem~1, the transformer simulates an iterative sampler by assigning transformer blocks to sampler steps. Therefore, depth is not only an architectural detail; it controls how many test-time computational steps the model can perform. If the model has too few layers, the transport process may be incomplete: the representation starts moving toward a reference geometry, but the model does not have enough intermediate computation to form a stable uniform-like phase before reconstructing semantic structure. We therefore interpret shallow or smaller models as a failure mode of in-context sampling.

\newpage

\subsection{Computational Resources}
All experiments were run on a single NVIDIA H100 GPU machine with 128 GB of system memory.  The controlled synthetic experiments in Appendix D.1 and D.2 are computationally lightweight: they use a small GPT-2-style decoder-only transformer. If training is mentioned, the model is trained for 2,000 optimization steps on 1,000 IID training sequences. For the pretrained-language-model analyses in Appendix D.3, no model fine-tuning is performed. 
The experiments require only forward passes through publicly available autoregressive language models in order to extract hidden states after each transformer block. 

\subsection{Existing Assets and Licenses}
\label{app:existing-assets}

We use publicly available pretrained language models only for evaluation and do not redistribute their weights. 
For each model family, we cite the original paper, technical report, or official release, and follow the license or model-card terms associated with the official release. 
The evaluated models include GPT-2~\cite{radford2019language}, Llama/Llama-3.3~\cite{touvron2023llama,grattafiori2024llama}, OpenLLaMA/OpenLM-Llama, Cerebras-GPT~\cite{dey2023cerebras}, Qwen2.5~\cite{qwen2024qwen2}, Falcon~\cite{almazrouei2023falcon}, OPT~\cite{zhang2022opt}, and BLOOM~\cite{workshop2022bloom}. 
The CBT dataset is cited as~\cite{hill2015goldilocks}. 
License and access terms are taken from the corresponding official repositories or model cards.

\begin{table}[h!]
\centering
\small
\caption{Existing pretrained models and datasets used in the experiments. We use these assets only for evaluation and do not redistribute model weights or dataset copies. License information is taken from the corresponding official release pages or model cards.}
\label{tab:existing-assets-licenses}
\begin{tabular}{p{0.20\linewidth} p{0.28\linewidth} p{0.24\linewidth} p{0.20\linewidth}}
\toprule
\textbf{Asset} & \textbf{Use in this paper} & \textbf{Source / citation} & \textbf{License or terms} \\
\midrule
GPT-2 
& Pretrained autoregressive language model for layerwise hidden-state analysis 
& OpenAI release; cited as~\cite{radford2019language} 
& MIT / Modified MIT license, according to the official model release or model card. \\

Llama / Llama-3.3-70B-Instruct 
& Pretrained autoregressive language model for semantic-topic and CBT experiments 
& Meta Llama releases; cited as~\cite{touvron2023llama,grattafiori2024llama} 
& Meta Llama community/model license; access and use governed by the official Llama license terms. \\

OpenLLaMA / OpenLM-Llama-13B 
& Pretrained autoregressive language model for semantic-topic experiments 
& OpenLLaMA/OpenLM Research official release or model card; cited as~\cite{openlm2023openllama}
& Apache-2.0 or permissive open-source license, according to the official model card. \\

Cerebras-GPT-13B 
& Pretrained autoregressive language model for semantic-topic and CBT experiments 
& Cerebras-GPT release; cited as~\cite{dey2023cerebras} 
& Apache-2.0 license. \\

Qwen2.5 family 
& Pretrained autoregressive language models for model-scale comparison 
& Qwen release; cited as~\cite{qwen2024qwen2} 
& Apache-2.0 for most open-source Qwen2.5 variants; some variants, including 3B and 72B, use model-specific Qwen license terms. \\

Falcon-40B 
& Pretrained autoregressive language model for additional model-family experiments 
& Falcon release; cited as~\cite{almazrouei2023falcon} 
& Apache-2.0 license, according to the official release/model card. \\

OPT-66B 
& Pretrained autoregressive language model for additional model-family experiments 
& OPT release; cited as~\cite{zhang2022opt} 
& Meta OPT license terms, according to the official model card or license file. \\

BLOOM-7B1 
& Pretrained autoregressive language model for additional model-family experiments 
& BigScience BLOOM release; cited as~\cite{workshop2022bloom} 
& BigScience RAIL License v1.0. \\

CBT dataset 
& Natural-text prompts for layerwise MMD experiments 
& Children's Book Test; cited as~\cite{hill2015goldilocks} 
& Dataset release terms associated with the original CBT release; the dataset is used only for evaluation. \\
\bottomrule
\end{tabular}
\end{table}


\end{document}